%% file: main.tex
\pdfoutput=1
\documentclass[letterpaper]{article}
\usepackage{iclr2027_conference,times}

\input{_settings}
\input{_macros}
\input{_metadata}

\iclrfinalcopy

\begin{document}
\maketitle
\lhead{}
\renewcommand{\headrulewidth}{0pt}
{\renewcommand{\thefootnote}{\Letter}\footnotetext{Corresponding author.}}

\input{sections/0_abs}

\input{sections/1_intro}

\input{sections/2_background}

\input{sections/3_method}

\input{sections/4_exp}

\input{sections/5_related}

\input{sections/6_conclusion}

\bibliographystyle{iclr2027_conference}
\bibliography{sit_cl/reference}

\ifincludeappendix
  \clearpage
  \appendix
  \input{sit_cl/appendix}

\fi

\end{document}

%% file: _settings.tex
\newif\ifincludeappendix
\includeappendixtrue

%% file: _macros.tex
\usepackage[T1]{fontenc}
\usepackage[utf8]{inputenc}
\usepackage{amsmath,amssymb,amsfonts,bm}
\usepackage{graphicx}
\usepackage{booktabs,multirow,threeparttable,tabularx}
\usepackage{nicefrac}
\usepackage{microtype}
\usepackage{subcaption}
\usepackage[table]{xcolor}
\usepackage{CJKutf8}
\usepackage[most]{tcolorbox}
\usepackage{enumitem}
\usepackage{wrapfig}
\usepackage{float}
\usepackage{placeins}
\usepackage{url}
\usepackage{marvosym}
\usepackage{hyperref}

\definecolor{citegreen}{HTML}{006B3D}
\hypersetup{
  colorlinks=true,
  citecolor=citegreen,
  linkcolor=red,
  urlcolor=black,
  filecolor=black,
  pdftitle={Unlocking the Potential of Contrastive Regularization for Diffusion Transformer Training},
  pdfauthor={Yi Tan, Beier Zhu, Sirui Chen, Yunzhe Lin, Yuan Zhou, Cheng Chen, Xingyu Zhu, Tianrun Chen, Yanbin Hao, Hanwang Zhang}
}

\newcommand{\ours}{\texttt{CRAN}}
\input{sit_cl/define}

%% file: sit_cl/define.tex
\def\eg{\emph{e.g.}}

\def\ie{\emph{i.e.}}

\definecolor{questionblue}{HTML}{2B8AC6}

\definecolor{steelblue}{RGB}{45, 125, 189}
\definecolor{mutedred}{RGB}{190, 72, 65}
\newcounter{finding}
\newcommand{\finding}[1]{%
    \refstepcounter{finding}%
    \begin{tcolorbox}[
        enhanced,
        breakable,
        colback=questionblue!6,
        colframe=questionblue!95,
        boxrule=0.75pt,
        arc=4pt,
        boxsep=3pt,
        left=4pt,
        right=4pt,
        top=4pt,
        bottom=4pt,
        before skip=8pt,
        after skip=8pt,
        drop shadow=gray!25!white
    ]
    \begin{center}
    #1
    \end{center}
    \end{tcolorbox}
}

%% file: _metadata.tex
\title{Unlocking the Potential of Contrastive Regularization for Diffusion Transformer Training}
\author{%
Yi Tan\textsuperscript{1},\ Beier Zhu\textsuperscript{2,\Letter},\ Sirui Chen\textsuperscript{3},\ Yunzhe Lin\textsuperscript{4},\ Yuan Zhou\textsuperscript{1},\\
\bf Cheng Chen\textsuperscript{1},\ Xingyu Zhu\textsuperscript{5},\ Tianrun Chen\textsuperscript{6},\ Yanbin Hao\textsuperscript{4},\ Hanwang Zhang\textsuperscript{1}\\[3pt]
\textsuperscript{1}NTU\quad \textsuperscript{2}USTC\quad \textsuperscript{3}Tongji University\quad \textsuperscript{4}HFUT\quad \textsuperscript{5}NUS\quad \textsuperscript{6}Moxin Technology\\[2pt]
\texttt{yi.tan@ntu.edu.sg}\quad \texttt{beier.zhu@ustc.edu.cn}
}

%% file: sections/0_abs.tex
\begin{abstract}
Representation regularization has emerged as an effective way to improve diffusion transformer training. Yet one of the most canonical regularizers--InfoNCE-style contrastive learning--remains largely underexplored in this setting. We naturally ask a question: can standard contrastive learning serve as an effective regularizer for diffusion transformers? Surprisingly, we find that naive InfoNCE regularization brings only marginal gains. Our analysis reveals that the objective is easily dominated by a shortcut: near-orthogonal noise components in diffusion inputs provide a bypass for enlarging negative-pair distances while weakening positive-pair alignment, especially at high noise levels.  To unlock the benefit of contrastive regularization, we propose \texttt{C}lean \texttt{R}epresentation-\texttt{AN}chored Contrastive (\ours) regularization. Specifically, \ours~anchors contrastive dispersion with clean-sample representation distances, transferring clean relational semantic structure to noisy diffusion representations to suppress the noise shortcut. Extensive experiments on ImageNet show that \ours~consistently improves diffusion transformer training across model scales and training budgets, outperforming internal feature-alignment and dispersive regularization baselines while matching external DINO-guided REPA. These results re-establish InfoNCE-based contrastive learning as an effective regularizer for diffusion training, once its noise-driven over-dispersion is properly constrained.

 \end{abstract}

%% file: sections/1_intro.tex
\section{Introduction}

Transformer-based diffusion and flow ~\citep{ho2020denoising,song2020score,albergo2023NormFlows,liuflow,lipmanflow} models now dominate generation tasks, from image synthesis~\citep{podell2023sdxl,saharia2022photorealistic,esser2024scaling} to video generation~\citep{polyak2024movie,brooks2024video}.
Recent studies have substantially improved diffusion transformer training through \textit{representation-level regularization}, including MAE-style auxiliary objectives, external feature alignment, internal self-representation alignment, and feature-dispersion regularization~\citep{zheng2023maskdit,zhu2024sddit,krause2025tread,yu2024representation,jiangrepresentation,haghighi2025layersync,wang2025diffuse}.
However, existing methods rely on specialized alignment or dispersion objectives, while the canonical contrastive learning objective, \ie, InfoNCE~\citep{he2020moco,chen2020simclr}, remains largely underexplored in diffusion transformer training. This is surprising because InfoNCE has been one of the most successful objectives for representation learning in discriminative vision tasks. We therefore ask a natural question:

\finding{
    \textit{Can standard InfoNCE contrastive learning serve as an effective representation regularizer for diffusion transformers?}
}

\input{tables/table_teaser_v1}

\noindent
Evidence from prior work and our empirical study \textbf{suggests a negative answer}.
Dispersive Loss~\citep{wang2025diffuse} (see their Table~2) shows that standard contrastive losses can degrade diffusion training with independently noised positive views, and require restricted noise gaps to obtain only limited gains.
Consistent with prior evidence, we further test whether InfoNCE can become effective under a stronger self-representation alignment setup.
We adopt the SRA framework~\citep{jiangrepresentation}, where a middle SiT representation is aligned to a deeper and less noisy self-representation, and extend this positive-only regularizer into a full InfoNCE objective by adding negative-pair discrimination.
However, as in Table~\ref{tab:fid-comparison}, InfoNCE brings almost no gain over SRA, reducing FID only from 28.7 to 28.5.

To understand this issue, we argue that vanilla InfoNCE suffers from a \textbf{noise shortcut}.
Diffusion inputs contain high-dimensional Gaussian noise, whose random components are known to be nearly orthogonal with high probability~\citep{cai2013distributions}.
Therefore, instead of learning semantic differences between images, the contrastive objective may simply exploit the naturally near-orthogonal noise components to enlarge negative-pair distances.
This provides an easy way to satisfy the repulsive regularization, but can lead to over-dispersion and weaken positive-pair alignment\footnote{Simply lowering the InfoNCE temperature does not remove this shortcut: Table~\ref{tab:fid-comparison-tau} shows that temperature tuning still fails to recover the expected benefit of InfoNCE.}.

To verify this argument, we conduct three diagnostic studies under standard InfoNCE regularization.
(1) We track the average cosine distance ($1-\mathrm{cos}(\mathbf{a},\mathbf{b})$) of positive-pairs and negative-pairs during training. As shown in Figure~\ref{fig:fig1}~(a), negative-pairs are rapidly pushed toward orthogonality, which we refer to as \textbf{over-dispersion}.
However, positive-pair distances are not properly minimized; instead, they first increase and \textit{remain larger than their initial value}, indicating that positive-pair alignment is \textbf{weakened} rather than strengthened.
(2) Figure~\ref{fig:fig1}~(b) shows that both positive- and negative-pair distances grow with noise strength, especially the negative-pair ones, suggesting that \textbf{stronger noise amplifies shortcut}-driven over-dispersion and weakened alignment. (3) We further consider a special case of positive-pairs: fixing the clean image while varying only the sampled noise, and measuring feature distances before and after the contrastive projection head. Figure~\ref{fig:fig1}~(c) shows that for the same image under different noise realizations, the contrastive projection head \textbf{amplifies noise-induced feature distances}, revealing that  it exploits noise variation as a shortcut even for positive pairs.

\input{figures/Teaser}

Motivated by the analyses above, we propose {\texttt{C}lean \texttt{R}epresentation-\texttt{AN}chored contrastive} (\ours) regularization.
The key idea is to prevent InfoNCE from over-dispersing noisy representations beyond the discriminative structure present in clean representations.
Specifically, \ours~anchors negative-pair dispersion in the contrastive objective using pairwise distances between clean-sample representations.
In this way, clean representation geometry guides noisy diffusion representations, clipping unnecessary negative repulsion once the clean anchor is reached and thereby shifting optimization toward positive-pair alignment.
As shown in Figure~\ref{fig:fig1}, \ours~mitigates the key failure modes of vanilla InfoNCE: it alleviates negative-pair over-dispersion and improves positive-pair alignment, reduces the sensitivity of representation distances to noise strength, and suppresses projection-head amplification of noise-induced distances.
Consistently, Table~\ref{tab:fid-comparison} shows that \ours~reduces FID from 28.5 to 26.3 over vanilla InfoNCE, demonstrating that \ours~effectively blocks the noise shortcut and unlocks the potential of
InfoNCE for diffusion transformer training.

Extensive experiments in Section~\ref{sec:exp} verify the effectiveness of \ours, which achieves significant improvements over feature-alignment methods using internal features.
Notably, using only internal feature contrastive learning, \ours~matches REPA~\cite{yu2024representation}, which relies on an external strong DINOv2 encoder \cite{oquab2024dinov2}, under the same training budget.

%% file: tables/table_teaser_v1.tex
\begin{table}[H]
\centering
\fontsize{8}{8}\selectfont
\setlength{\tabcolsep}{10pt}

\begin{minipage}[c]{0.5\linewidth}
\centering
\begin{tabular}{llc}
\toprule
\textbf{Method} &
\textbf{Regularization} &
\textbf{FID ($\downarrow$)} \\
\midrule
Baseline & None & 33.0 \\
SRA & Positive only & 28.7 \\
InfoNCE & Positive + Negative & 28.5 \\
\midrule
\ours~(ours) & Anchored InfoNCE & \textbf{26.3} \\
\bottomrule
\end{tabular}
\end{minipage}
\hfill
\begin{minipage}[c]{0.45\linewidth}
\caption{
FID-50K of SiT-B/2 trained on ImageNet 256$\times$256 for 400K iterations.
Standard InfoNCE regularization brings only marginal gains over the positive-only
representation regularizer SRA~\citep{jiangrepresentation}.
}
\label{tab:fid-comparison}
\end{minipage}

\vspace{-4mm}
\end{table}

%% file: figures/Teaser.tex
\begin{figure*}[t]
\centering
\includegraphics[width=\linewidth]{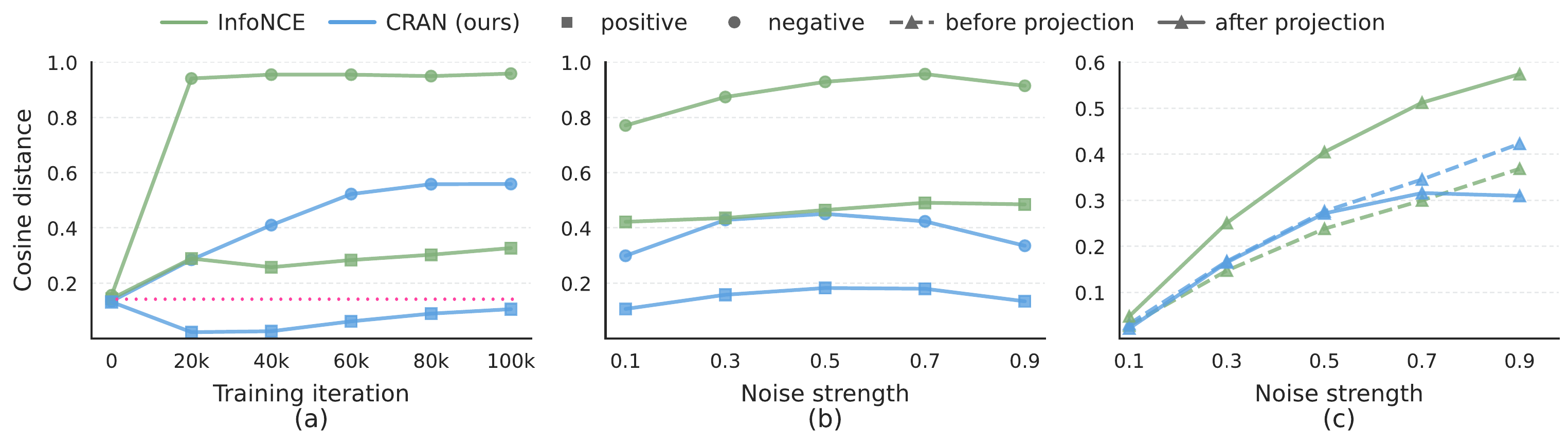}
\vspace{-7mm}
\caption{\textbf{Diagnostic studies of the noise shortcut.}
(a) Standard InfoNCE rapidly over-disperses negative pairs and weakens positive-pair alignment, while \ours~alleviates both effects.
(b) The over-dispersion and weakened alignment of InfoNCE become more severe at higher noise levels, whereas \ours~remains stable.
(c) For a fixed clean image, the contrastive projection head amplifies noise-induced distances, revealing a noise shortcut; \ours~suppresses such amplification.}

\label{fig:fig1}
\vspace{-3mm}
\end{figure*}

%% file: sections/2_background.tex
\section{Preliminary}
\vspace{-1mm}

\subsection{Flow and diffusion models}
We use the stochastic interpolant formulation~\citep{albergo2025stochastic} as a unified perspective for flow and diffusion models.
Let $\mathbf{x}_{0} \sim p_{\mathsf{data}}$ denote an image latent and $\bm{\epsilon} \sim \mathcal{N}(\mathbf{0}, I)$ denote Gaussian noise.
The interpolated state at timestep $t$ is defined as $\mathbf{x}_{t}=\alpha_t\mathbf{x}_{0}+\sigma_t\bm{\epsilon}$, where $\alpha_t=\alpha(t)$ and $\sigma_t=\sigma(t)$ are time-dependent coefficients, with $\alpha(t)$ decreasing and $\sigma(t)$ increasing over time.
The probability-flow ODE is governed by the velocity field:
$
\dot{\mathbf{x}}_{t}
=
\mathbf{v}(\mathbf{x}_{t}, t)
$.
In practice, $\mathbf{v}_{\bm{\theta}}$ is trained to regress the sample-wise velocity target $\dot{\alpha}_{t}\mathbf{x}_{0}+\dot{\sigma}_{t}\bm{\epsilon}$:
\begin{equation}
\mathcal{L}_{\mathsf{gen}}(\bm{\theta})
=
\mathbb{E}_{\mathbf{x}_{0}, \bm{\epsilon}, t}
\left[
\left\|
\mathbf{v}_{\bm{\theta}}(\mathbf{x}_{t}, t)
- \left(\dot{\alpha}_{t}\mathbf{x}_{0}
+ \dot{\sigma}_{t}\bm{\epsilon}\right)
\right\|^{2}_2
\right].
\label{eq:velocity_loss}
\end{equation}
Here, $\dot{\alpha}_{t}$ and
$\dot{\sigma}_{t}$ denote time derivatives.
Different choices of $\alpha_t$ and $\sigma_t$ recover common flow and diffusion
parameterizations. For example, flow model uses
$\alpha_t=1-t$ and $\sigma_t=t$, giving the target
$\bm{\epsilon}-\mathbf{x}_0$; In VE diffusion~\citep{song2020score}, $\alpha_t=1$ and $\sigma_t=\sigma(t)$, yielding a
pure noise-direction target $\dot{\sigma}_t\bm{\epsilon}$.

\input{tables/unified_regularization}

\subsection{Existing representation regularizers for diffusion training}

Recent studies show that representation regularization is effective for diffusion training.
A unified view of diffusion representation regularization is summarized in Table~\ref{tab:rep_reg_unified}.
Based on how representation pairs are constructed and optimized, we categorize existing methods into three families.

\noindent\textbf{Positive-only regularization.} Positive-only regularizers improve diffusion training by aligning each noisy intermediate representation with a stronger target feature.
Representative methods differ mainly in how this target feature is constructed:

\begin{itemize}[leftmargin=1em,labelsep=0.3em,itemsep=0.em]
    \item {REPA}~\citep{yu2024representation} minimizes the cosine distance between intermediate diffusion features and representations from an external pretrained vision encoder, \eg, DINOv2~\citep{oquab2024dinov2}.

    \item LayerSync~\citep{haghighi2025layersync} directly minimizes the cosine distance between intermediate diffusion representations and detached deeper representations within the same diffusion transformer.

    \item SRA~\citep{jiangrepresentation} also aligns intermediate representations with detached deeper representations, but constructs the target from an EMA teacher network  and a less noisy version of the current latent, using an $\ell_2$ distance objective.

    \item {SRA2}~\citep{wang2026SRA2} uses intermediate diffusion representations to predict the VAE latent of the clean image.
\end{itemize}

\noindent\textbf{Negative-only regularization.}
Negative-only regularizer disperses intermediate diffusion representations across samples without positive pairs.
Dispersive Loss~\citep{wang2025diffuse} instantiates this idea by minimizing a log-sum-exp objective over negative pairwise $\ell_2^2$ distances.

\noindent\textbf{Positive-negative contrastive regularization.}
Contrastive learning is one of the most successful objectives for visual representation learning~\citep{chen2020simclr,he2020moco,radford2021learning}.
It learns discriminative features by aligning positive pairs while repelling negative pairs.
Let $\mathbf{z}_i$ denote the query representation, $\mathbf{y}_i$ its positive target, and $\{\mathbf{y}_j\}_{j\neq i}$ the in-batch negatives.
Given a distance function $d(\cdot,\cdot)$, the InfoNCE-based contrastive objective is
\begin{equation}
\mathcal{L}_{\mathsf{CL}}(\bm{\theta},\bm{\psi})
=
-\mathbb{E}_i\left[
\log
\frac{
\exp\!\left(-d(\mathbf{z}^{i}, \mathbf{y}^{i})/\tau\right)
}{
\sum_{j=1}^{B}
\exp\!\left(-d(\mathbf{z}^{i}, \mathbf{y}^{j})/\tau\right)
}\right].
\label{eq:contrastive_loss}
\end{equation}
However, directly applying this full objective to diffusion training remains challenging.
Dispersive Loss~\citep{wang2025diffuse} shows that $\mathcal{L}_\mathsf{CL}$ can degrade generation quality, and only yields limited gains after carefully restricting the noise gap between positive views.
We revisit this problem  and find that vanilla InfoNCE is dominated by noise-driven dispersion rather than semantic discrimination. Moreover, this shortcut cannot be resolved by simply tuning the temperature $\tau$: as shown in Table~\ref{tab:fid-comparison-tau}, using a smaller temperature does not recover the expected benefit of InfoNCE.

%% file: tables/unified_regularization.tex
\begin{table*}[t]
\centering
\caption{
\textbf{Unified view of representation regularization methods for diffusion training.}
\textit{P}: positive;
\textit{N}: negative;
$\mathbf{x}_0^i/\mathbf{x}_t^i/\mathbf{x}_{t-\Delta t}^i$: clean/noisy/less-noisy query image;
$\mathbf{z}_t^i=g_{\bm{\psi}}(f_{\bm{\theta}}^k(\mathbf{x}_t^i))$: $k$-th block feature of noisy query projected by head $g_{\bm{\psi}}$.
$\mathrm{sg}$: stop-gradient; $d(\cdot,\cdot)$: distance function;
$f_{\bar{\bm{\theta}}}$: EMA model; $f^{k^\prime}$ with $k^\prime > k$: deeper-layer block; $h$: auxiliary reconstruction head.
To ease the notation, we omit the temperature $\tau$ in Dispersive Loss~\citep{wang2025diffuse}, InfoNCE, and \ours. $^\dagger$ indicates the use of external knowledge.
}
\label{tab:rep_reg_unified}
\scriptsize
\setlength{\tabcolsep}{3.5pt}
\begin{tabular}{ll p{0.33\linewidth}l}
\toprule
\textbf{Type} & \textbf{Method} & \textbf{Target / Pair Construction} & \textbf{Regularization Objective} \\
\midrule

\multirow{4}{*}[-6ex]{\textit{P.-only}}
& REPA$^\dag$~\citep{yu2024representation}
& External pretrained representation $\mathbf{y}^i=E_\mathsf{DINO}(\mathbf{x}^{i}_{0})$
& $\mathbb{E}_i\left[ d\!\left(\mathbf{z}^{i}_{t}, \mathrm{sg}(\mathbf{y}^i)\right)\right]$ \\

& LayerSync~\citep{haghighi2025layersync}
& Detached deeper-layer representation
$\mathbf{y}^i=f_{\bm{\theta}}^{k'}(\mathbf{x}^{i}_{t})$
& $\mathbb{E}_i\left[
d\!\left(\mathbf{z}^{i}_{t}, \mathrm{sg}(\mathbf{y}^i)\right)\right]$ \\

& SRA~\citep{jiangrepresentation}
& Deeper and less noisy EMA representation
$\mathbf{y}^i=f_{\bar{\bm{\theta}}}^{k'}(\mathbf{x}^{i}_{t-\Delta t})$
& $\mathbb{E}_i
\left[\|\mathbf{z}^{i}_{t}-\mathrm{sg}(\mathbf{y}^i)\|_2^2\right]$ \\

& SRA2~\citep{wang2026SRA2}
& Reconstructing VAE latent of clean image $\mathbf{y}^i=\mathbf{x}^{i}_{0}$
& $\mathbb{E}_i \left[
\left\|h(\mathbf{z}^{i}_{t})-\mathrm{sg}(\mathbf{y}^{i})\right\|_2^2 \right]$ \\

\midrule

\textit{N.-only}
& Disp. Loss~\citep{wang2025diffuse}
& In-batch negatives $\{\mathbf{y}^j=\mathbf{z}_t^j\}_{j\ne i}$
& $\log \mathbb{E}_{i,j}
\left[\exp\left(-\|\mathbf{z}^{i}_{t}-\mathbf{y}^{j}\|_2^2\right)\right]$ \\

\midrule
\multirow{2}{*}[-3.5ex]{\textit{P. + N.}}
& InfoNCE
& Positive $\mathbf{y}^i$ and in-batch negatives $\{\mathbf{y}^j\}_{j\ne i}$
& $
 -\mathbb{E}_{i}\left[
\log
\frac{
\exp(-d(\mathbf{z}^{i}_{t},\mathbf{y}^i))
}{
\sum_j \exp(-d(\mathbf{z}^{i}_{t},\mathbf{y}^j))
}\right]$ \\

& \ours~(ours)
& InfoNCE + clean-representation anchor
${\color{mutedred}m}
=\mathbb{E}_{i,j}[d(\mathbf{z}^{i}_{0},\mathbf{z}^{j}_{0})]$
& $
 -\mathbb{E}_{i}\left[
\log
\frac{
\exp(-d(\mathbf{z}^{i}_{t},\mathbf{y}^i))
}{
\sum_j \exp(-\min(d(\mathbf{z}^{i}_{t},\mathbf{y}^j)
,{\color{mutedred}m}
)}\right]$ \\

\bottomrule
\end{tabular}
\vspace{-1mm}
\end{table*}

%% file: sections/3_method.tex
\section{Methodology}

Given a noisy query latent  $\mathbf{x}_{t}^{i}$, let
$\mathbf{z}^{i}_{t}=g_{\bm{\psi}}(f_{\bm{\theta}}^{k}(\mathbf{x}_{t}^{i}))$
denote the projected intermediate diffusion representation, where
$f_{\bm{\theta}}^{k}(\cdot)$ produces the $k$-th block features and $g_{\bm{\psi}}(\cdot)$ is an MLP projection head parameterized by $\bm{\psi}$.
Let $\mathbf{y}^{i}$ denotes the detached positive or negative representations, \eg, a deeper and less noisy self-representation as in SRA~\citep{jiangrepresentation}:
\begin{equation}
    \mathbf{y}^i=f_{\bar{\bm{\theta}}}^{k^\prime}(\mathbf{x}_{t-\Delta t}^i), \ \text{where}\ k^\prime > k \ \text{and}\ 0 < t- \Delta t < t.
\end{equation}
Here, $\bar{\bm{\theta}}$ denotes the EMA parameters.
For convenience, we define
\[
d^{ij}_{t}=d(\mathbf{z}^{i}_{t}, \mathbf{y}^{j}),
\]
where $d(\cdot,\cdot)$ denotes a representation distance, such as squared $\ell_2$ distance or cosine distance.

\subsection{Fixed-anchor contrastive regularization}

Evidence from prior work~\citep{wang2025diffuse} and our empirical analysis in Table~\ref{tab:fid-comparison} suggest that standard InfoNCE in Eq.~\eqref{eq:contrastive_loss} is not directly effective for diffusion training.
We attribute this failure to noise-shortcut-induced negative-pair over-dispersion.
Since diffusion representations contain high-dimensional noise components that are naturally near-orthogonal, contrastive learning can satisfy its repulsive objective through noise-induced separation rather than semantic discrimination, as shown in Figure~\ref{fig:fig1}.

A direct remedy is therefore to stop the contrastive objective from further pushing
negative pairs that are already sufficiently separated.
To this end, we introduce Anchored Contrastive Learning (ACL) with a fixed distance anchor $m_{\mathsf{fix}}$ and clip the
negative-pair distances in the InfoNCE denominator:
 \begin{equation}
\mathcal{L}_{\mathsf{ACL}}(\bm{\theta},{\bm{\psi}},m_\mathsf{fix})
=
-\mathbb{E}_i \left[
\log
\frac{
\exp(-d_{t}^{ii}/\tau)
}{
\sum_{j}
\exp\!\left(
-\min(d_{t}^{ij}, m_{\mathsf{fix}})/\tau
\right)
}\right].
\label{eq:fixed_anchor_loss}
\end{equation}
 Here, $\min(d_{t}^{ij}, m_{\mathsf{fix}})$ limits the maximum separation enforced on negative pairs. Thus, the anchor prevents InfoNCE from blindly pushing negative pairs toward orthogonality.
Once a negative pair reaches the distance bound, it no longer receives additional repulsive gradients, so the loss can be further reduced mainly by improving positive-pair alignment.
This both alleviates over-dispersion and strengthens the alignment of matched representations.

Notably, our results in Table~\ref{tab:fid-comparison-tau} show that a fixed anchor already validates this intuition: setting $m_{\mathsf{fix}}=0.5$ ($\cos 60^\circ$) improves FID from 28.5 to 27.0 on SiT-B/2.

\begin{figure}[t]
  \centering
  \includegraphics[width=\linewidth]{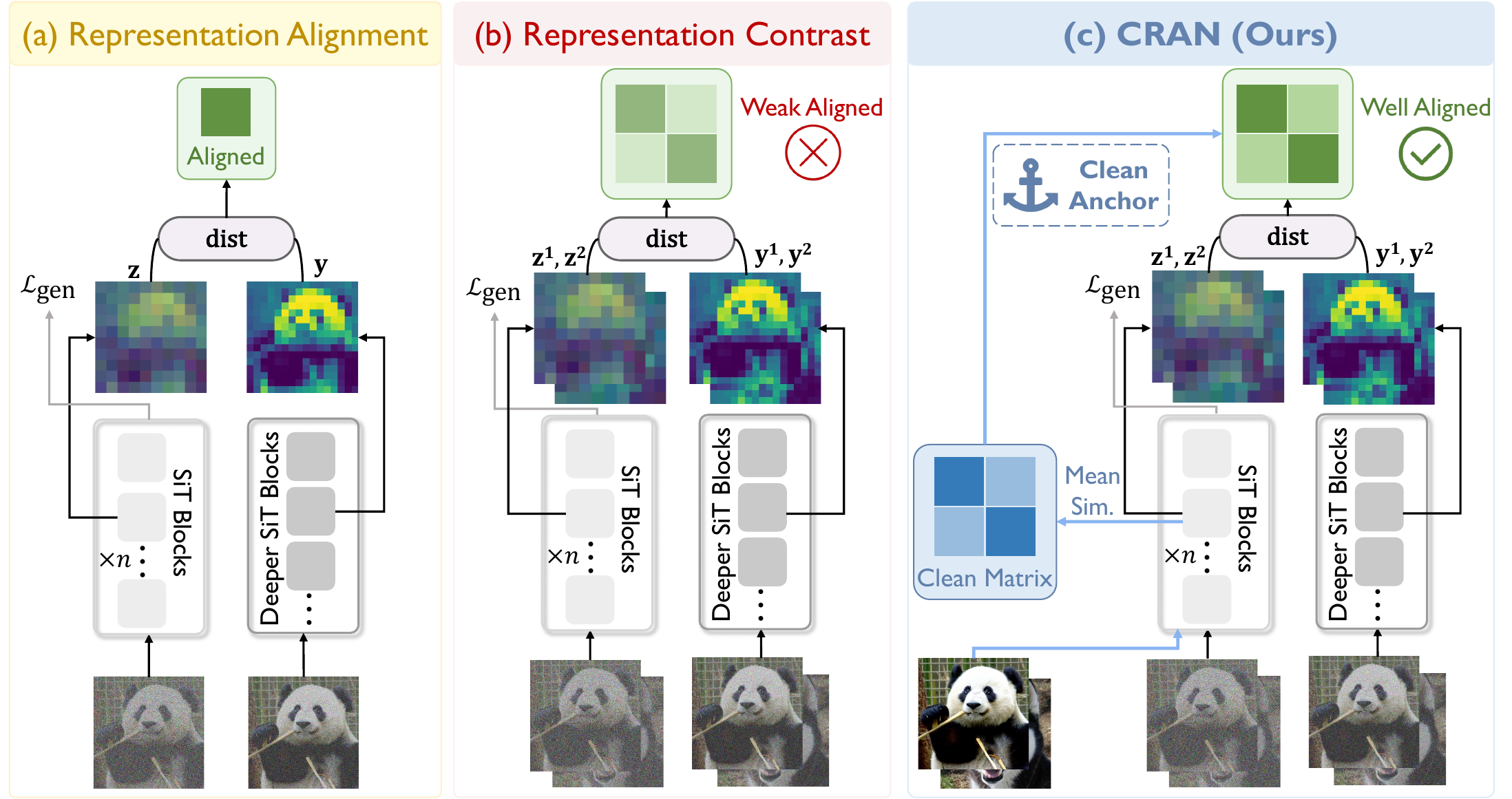}
  \caption{
  \textbf{(a)} Prior diffusion representation alignment methods, \eg, SRA~\citep{jiangrepresentation}.
  \textbf{(b)} Standard contrastive learning leads to severe over-dispersion and weakened alignment, resulting in limited performance gains.
  \textbf{(c)} \ours~uses clean-representation distances to anchor the maximum contrastive separation, alleviating over-dispersion and improving positive-pair alignment, which yields substantially stronger performance than representation alignment.
  }
  \vspace{0mm}
  \label{fig:method}
\end{figure}

\subsection{{Clean representation-anchored contrastive}  regularization}

Although the fixed-anchor variant validates the importance of limiting negative-pair dispersion, it remains suboptimal for two reasons:
{(1)} it requires manually tuning an extra hyperparameter;
{(2)} its fixed threshold cannot adapt to the evolving inter-representation geometry during training.
To remove this hand-crafted design, we turn to clean-sample representations.
Unlike noisy diffusion representations, clean representations are not dominated by injected Gaussian noise and thus provide a more reliable estimate of semantic representation geometry.

Motivated by this, we propose {\texttt{C}lean \texttt{R}epresentation-\texttt{AN}chored contrastive} (\ours) regularization, as in Figure~\ref{fig:method}.
Instead of using a fixed anchor, \ours~uses the average negative pairwise distance among clean-sample representations as an adaptive anchor ($m_\mathsf{cran}$) for noisy contrastive learning.
\begin{equation}
\label{eq:cran_loss}
\mathcal{L}_{\mathsf{CRAN}}(\bm{\theta},{\bm{\psi}})
= \mathcal{L}_\mathsf{ACL}(\bm{\theta}, {\bm{\psi}}, m_\mathsf{cran}),\ \text{where}\
m_\mathsf{cran} = \mathbb{E}_{i\neq j}\left[d(\mathbf{z}_0^i,\mathbf{z}_0^j)\right]
.
\end{equation}
Here, $\mathbf{z}_0^i=g_{\psi}(f_{\bm{\theta}}^{k}(\mathbf{x}_{0}^{i}))$ denotes the projected representation of the clean input, and $m_\mathsf{cran}$ is computed as the empirical mean within the current batch.
This allows noisy contrastive learning to adaptively inherit the clean inter-sample geometry. The overall objective is given by:
\begin{equation}
\mathcal{L}_{\mathsf{total}}(\bm{\theta},{\bm{\psi}})
=
\mathcal{L}_{\mathsf{gen}}(\bm\theta)
+
\lambda \mathcal{L}_{\mathsf{CRAN}}(\bm{\theta},{\bm{\psi}}),
\label{eq:total_loss}
\end{equation}
where $\mathcal{L}_\mathsf{gen}$ is the flow or diffusion loss defined in Eq.~\eqref{eq:velocity_loss} and $\lambda$ controls the strength of regularization.

\vspace{-1mm}
\subsection{Clean Inter-Sample Geometry Estimation}
During training, each batch naturally contains samples at different noise levels. We select samples whose noise levels are below a threshold $t_c=0.1$ and use the average pairwise angle between their representations to determine the adaptive margin $m_{\mathrm{cran}}$. In this way, $m_{\mathrm{cran}}$ provides an online estimate of the clean inter-sample geometry without requiring an additional clean-sample branch.

%% file: sections/4_exp.tex
\definecolor{lightblue}{RGB}{235,248,255}
\section{Experiments}\label{sec:exp}
In this section, we conduct extensive experiments to answer the following research questions (RQs):

\noindent\textbf{RQ1:} Can \ours~significantly improve the performance of diffusion transformers and  surpass current state-of-the-art representation regularization methods?

\noindent\textbf{RQ2:} How do \ours-specific choices $(d,\lambda)$ affect performance, and can temperature tuning $\tau$ in naive InfoNCE or fixed-anchor contrastive regularization match the performance of our \ours?

\noindent\textbf{RQ3:} How does the adaptive anchor $m_\mathsf{cran}$ evolve in SiT+\ours~training?

\noindent\textbf{RQ4:} What representation properties are improved by \ours?

\input{tables/main_results}

\textbf{Implementation details.}
We conduct experiments on Scalable Interpolant Transformers (SiT)~\citep{ma2024sit}, which provide a unified view of diffusion and flow models.
Our evaluation covers three model scales: SiT-B/2, SiT-L/2, and SiT-XL/2.
For a fair comparison, we follow the REPA~\citep{yu2024representation} training configuration.
All experiments are carried out on ImageNet~\citep{deng2009imagenet}, with images center-cropped and resized to $256 \times 256$.
They are then encoded by the Stable Diffusion VAE with a compression factor of $8$, resulting in latents with shape $\mathbb{R}^{32 \times 32 \times 4}$.
Training is performed with a batch size of $256$ using AdamW, a constant learning rate of $1 \times 10^{-4}$, and no weight decay.

\textbf{Inference and evaluation.}
For inference, we follow SiT and REPA and use the Euler-Maruyama sampler for the SDE with $w_t=\sigma_t$, with the default number of function evaluations (NFE) set to $250$.
For evaluation, we adopt the ADM TensorFlow evaluation protocol~\citep{dhariwal2021diffusion}, generating $50$K samples to ensure statistical reliability and comparing them against the corresponding reference statistics.
We report Fr\'echet Inception Distance (FID)~\citep{heusel2017gans}, structural FID (sFID)~\citep{nash2021sparse}, Inception Score (IS)~\citep{salimans2016improved}, precision, and recall~\citep{kynkaanniemi2019improved} to assess generation quality from complementary perspectives.
Additional implementation details are provided in the Appendix \ref{appendix:Experimental setups}.

\subsection{The effectiveness of CRAN (RQ1)}

We first compare the improvements brought by \ours~on SiT across different model scales with those of other representation alignment methods. We use FID as the primary comparison metric and do not use classifier-free guidance. As shown in Table~\ref{tab:rq1-fid-nocfg}, \ours~substantially improves SiT across different model sizes and training budgets. Compared with existing representation alignment methods, including positive regularizers such as SRA~\citep{jiangrepresentation} and SRA2~\citep{wang2026SRA2}, as well as negative regularizers such as Dispersive Loss~\citep{wang2025diffuse} and DiverseDiT~\citep{yang2026diversedit}, \ours~achieves the best performance across different training stages and model sizes. It is worth noting that simply combining SRA with Dispersive Loss yields only marginal improvement and remains notably behind \ours~(28.6 vs. 26.3). Specifically, \ours~achieves 7.5 non-cfg FID using only 1M iterations, which outperforms vanilla SiT-XL/2 trained for 7M iterations. The performance of \ours continues to improve with more training iteration, and finally achieves 6.2 FID, which is the best result among methods that do not use external pretrained visual encoders.

\vspace{1mm}
Finally, we use SiT-XL/2 as the baseline and compare \ours~with other state-of-the-art methods under classifier-free guidance. As shown in Table~\ref{tab:rq1-main-comparison}, \ours~substantially improves the performance of SiT and reaches an FID of 1.45 at 800 epochs, outperforming MaskDiT~\citep{zheng2023maskdit} and SD-DiT~\citep{zhu2024sddit}, which introduce more complex architectures. Compared with representation alignment methods built on the same SiT-XL/2 backbone, \ours~achieves the best FID among methods that rely on features from the diffusion transformer itself, surpassing SRA~\citep{jiangrepresentation}, SRA2~\citep{wang2026SRA2} and LayerSync~\citep{haghighi2025layersync}.

\begin{figure}[t]
  \centering
  \includegraphics[width=\linewidth]{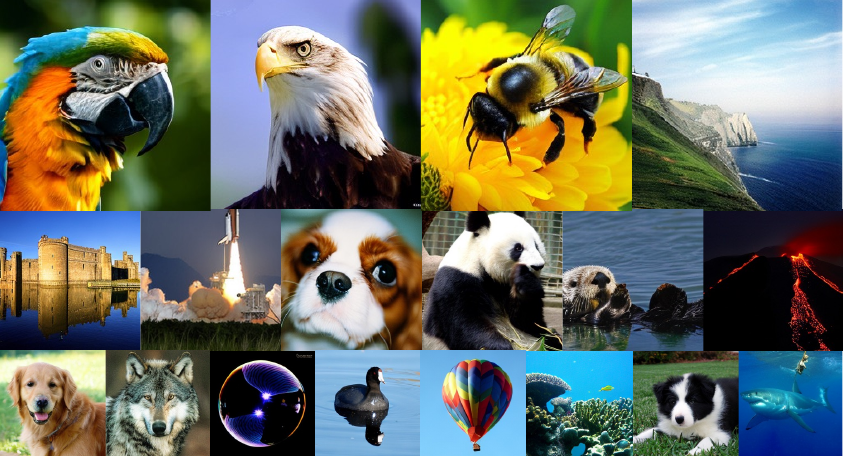}
  \caption{Generated images from SiT-XL/2 + \ours. More visualizations are provided in Appendix~\ref{appendix:Samples}.}

\end{figure}

\subsection{Ablation studies (RQ2)}

\begin{table}[t]
\centering
\fontsize{8}{10}\selectfont
\setlength{\tabcolsep}{10pt}
\vspace{-1mm}
\caption{Ablation studies on the design choices of \ours. We study the distance metric, the setting of the maximum separation, and the weight of the \ours~objective. The best results are in blue.}
\label{tab:ablation}
\begin{tabular}{lcccccccc}
\toprule
Method & dis. metric & Dis. Anchor & $\lambda$ & FID$\downarrow$ & sFID$\downarrow$ & IS$\uparrow$ & Pre.$\uparrow$ & Rec.$\uparrow$ \\
\midrule
SiT-B/2 & - & - & - & 33.0 & 6.46 & 43.7 & 0.53 & 0.63 \\
\midrule
\quad + \ours & \cellcolor{gray!10} $\ell_2$ & - & 1.0 & 31.9 & 6.53 & 44.8 & 0.55 & 0.63 \\
\quad + \ours & \cellcolor{gray!10} $\cos$ & - & 1.0 & \cellcolor{lightblue}28.3 & \cellcolor{lightblue}5.92 & \cellcolor{lightblue}50.9 & \cellcolor{lightblue}0.57 & \cellcolor{lightblue}0.63 \\
\midrule
\quad + \ours & $\cos$ & \cellcolor{gray!10}$\cos 30^\circ$ & 1.0 & 27.9 & 5.99 & 51.3 & 0.57 & 0.64 \\
\quad + \ours & $\cos$ & \cellcolor{gray!10}$\cos 60^\circ$ & 1.0 & 27.0 & 6.06 & 53.9 & 0.58 & 0.64 \\
\quad + \ours & $\cos$ & \cellcolor{gray!10}$\cos 70^\circ$ & 1.0 & 28.6 & 5.99 & 50.5 & 0.57 & 0.64 \\
\quad + \ours & $\cos$ & \cellcolor{gray!10}$m_\mathsf{cran}$ & 1.0 & \cellcolor{lightblue}26.3 & \cellcolor{lightblue}6.16 & \cellcolor{lightblue}56.2 & \cellcolor{lightblue}0.58 & \cellcolor{lightblue}0.63 \\
\midrule
\quad + \ours & $\cos$ & $m_\mathsf{cran}$ & \cellcolor{gray!10}0.5 & 28.6 & 6.08 & 50.6 & 0.57 & 0.63 \\
\quad + \ours & $\cos$ & $m_\mathsf{cran}$ & \cellcolor{gray!10}1.0 &  \cellcolor{lightblue}26.3 & \cellcolor{lightblue}6.16 & \cellcolor{lightblue}56.2 & \cellcolor{lightblue}0.58 & \cellcolor{lightblue}0.63 \\
\quad + \ours & $\cos$ & $m_\mathsf{cran}$ & \cellcolor{gray!10}1.5 & 26.8 & 6.12 & 54.3 & 0.58 & 0.63 \\
\bottomrule
\end{tabular}
\vspace{0mm}
\end{table}

As shown in Table~\ref{tab:ablation}, we study the key design choices in \ours, including the distance metric, the setting of the similarity anchor, and the weight of the \ours~regularization. We first evaluate representation contrastive regularization with different distance metrics. The results show that plain representation contrastive regularization can improve SiT, but the gains remain relatively marginal.
Between the $\ell_2$ distance and the cosine distance, we choose the cosine distance as our distance metric, as it achieves better performance and is widely used in representation learning to constrain the angular relationships between features~\citep{he2020moco,radford2021learning,chen2020simclr}.

\vspace{1mm plus 1mm}
Next, we study how different similarity anchor settings affect performance. A fixed threshold can improve performance when properly tuned. However, using the similarity between clean features as the anchor value achieves better results. We attribute this to the fact that the clean representation similarity more faithfully reflects the relationships among images at the representation level, while also allowing the hinge value to adapt dynamically during training.

\vspace{1mm plus 1mm}
We then study the effect of the \ours~weight. The results show that too small a weight degrades the performance of \ours. We therefore use $\lambda=1.0$ as the default weight in our final setting.

\vspace{0mm plus 1mm}
\begin{wraptable}{r}{0.40\linewidth}
  \vspace{-1.0em}
  \centering
  \small
  \setlength{\tabcolsep}{5pt}
  \caption{Component ablation on SiT-B/2}
  \label{tab:component-ablation}
  \begin{tabular}{lccc}
  \toprule
  Method & Pos. & Neg. & FID$\downarrow$ \\
  \midrule
  SiT-B/2 & No & No & 33.0 \\
  \ours & No & Yes & 31.9 \\
  \ours & Yes & No & 28.7 \\
  \ours & Yes & Yes & \textbf{26.3} \\
  \bottomrule
  \end{tabular}
  \vspace{-0.8em}
\end{wraptable}
Table~\ref{tab:component-ablation} isolates the two components of the contrastive objective. Removing either the positive or negative component degrades performance. The positive term provides cross-layer and cross-noise alignment, while the anchored negative term preserves useful inter-sample discrimination; their combination yields the strongest overall result. We further compare temperature scaling, fixed anchors, and the adaptive clean anchor in Appendix~\ref{apendix:More ablation studies and analysis} (Table~\ref{tab:fid-comparison-tau}).

\WFclear
\vspace{1mm}
\subsection{Dynamics of clean representation anchor (RQ3)}
We visualize the evolution of the clean representation anchor $m_\mathsf{cran}$ during training on SiT-B/2, SiT-L/2, and SiT-XL/2.\par
\begin{wrapfigure}{r}{0.50\linewidth}
  \vspace{-1.1em}
  \centering
  \includegraphics[width=\linewidth]{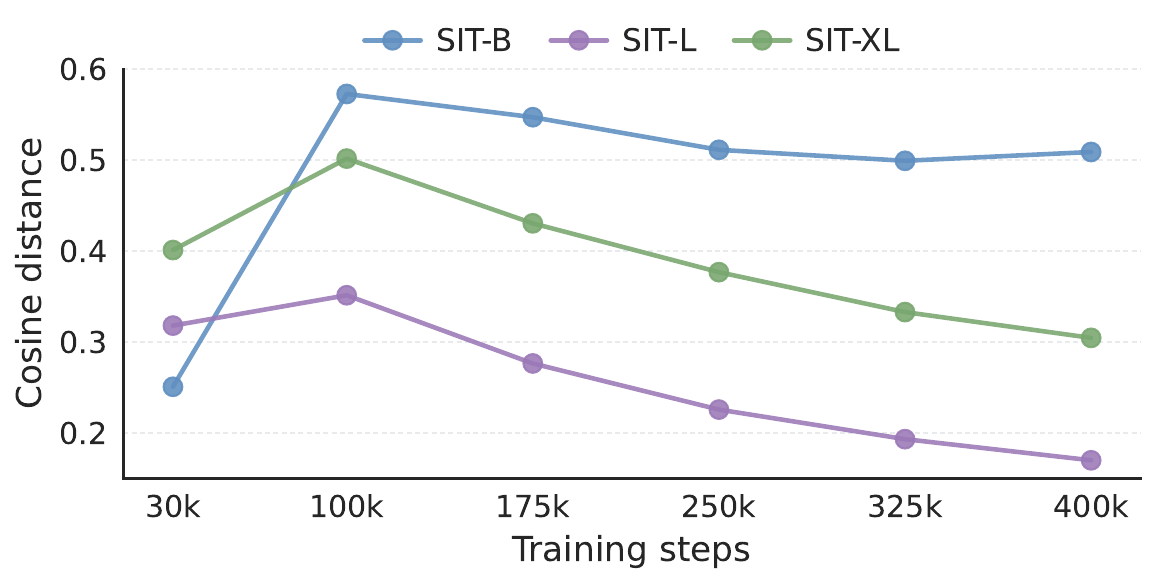}
  \caption{Evolution of the clean representation anchor during training.}
  \label{fig:angle_train_step}
  \vspace{-1.0em}
\end{wrapfigure}
As shown in Fig.~\ref{fig:angle_train_step}, $m_\mathsf{cran}$ follows a consistent pattern across model scales.
It increases rapidly in the early training stage, providing stronger discriminative regularization when the diffusion representations are still underdeveloped.
After reaching its peak, the anchor gradually decreases and eventually stabilizes as the model becomes more stable.
Among the three models, SiT-B/2 maintains the largest clean representation anchor.
This may be because its smaller capacity requires a more uniformly distributed representation space, whereas larger models can rely on weaker regularization after convergence.

\vspace{1mm}
\subsection{Representation analysis (RQ4)}
\label{sec:representation-analysis}
\noindent\textbf{Spatial representation structure.}
To examine how \ours~affects diffusion representations, we follow the iREPA~\citep{singh2025whatmatters} evaluation paradigm, which emphasizes that spatial-structure metrics are more closely aligned with diffusion generative performance than linear probing accuracy.
Specifically, we evaluate the learned representations using three metrics: Local vs. Distant Similarity (LDS)~\citep{huang1997image}, which measures the contrast between local and distant patch similarities; Correlation Decay Slope (CDS), which quantifies how quickly patch similarity decreases with spatial distance; and RMS Spatial Contrast (RMSC), which measures the spatial variation of normalized patch features.
These metrics allow us to quantitatively compare the representations learned by \ours~against those of competing methods.
The detailed computation procedures for these metrics are provided in the Appendix~\ref{appendix:metrics}.

\begin{figure}[t]
  \centering
  \includegraphics[width=\linewidth]{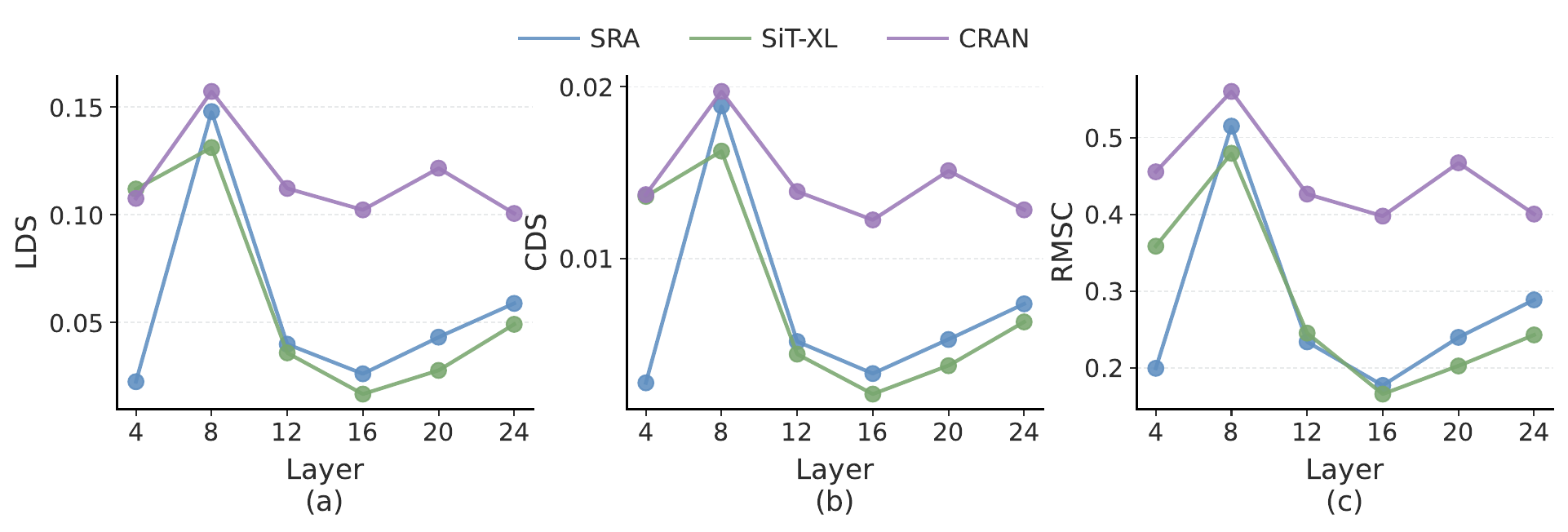}
  \caption{Spatial-structure metric analysis using SiT-XL/2. \ours~outperforms both SiT and SRA at the aligned layer, with even more pronounced improvements observed in deeper layers.}
    \vspace{-3mm}
  \label{fig:spatial_metrics}
\end{figure}

\vspace{0mm plus 1mm}
We compare our \ours~with a representative representation-alignment method, \ie, SRA~\citep{jiangrepresentation} which aligns representation across layers and time steps. As shown in Fig.~\ref{fig:spatial_metrics}, \ours demonstrates more significant enhancement in all three spatial-structure metrics at the regularized layer (layer 8). At deeper layers, \ours~ brings more pronounced improvements.

%% file: tables/main_results.tex
\begin{table}[t]
\centering
\begin{minipage}[t]{0.43\linewidth}
  \vspace{0pt}
  \fontsize{8}{9}\selectfont
  \setlength{\tabcolsep}{2.4pt}
  \renewcommand{\arraystretch}{0.82}
  \centering
  \captionof{table}{FID comparison across model scales and training budgets without CFG.}
  \label{tab:rq1-fid-nocfg}
  \begin{tabular}{lccc}
  \toprule
  Model & \#Params & Iter. & FID$\downarrow$ \\
  \midrule
  SiT-B/2 & 130M & 400k & 33.0 \\
  \quad + SRA & 130M & 400k & 29.1 \\
  \quad + SRA \& Disp. & 130M & 400k & 28.6 \\
  \quad + DiverseDiT & >130M & 400k & 28.1 \\
  \quad + SRA2 & 130M & 400k & 28.9 \\
  \rowcolor{lightblue}\quad + \ours & 130M & 400k & \textbf{26.3} \\
  \midrule
  SiT-L/2 & 458M & 400k & 18.8 \\
  \quad + SRA & 458M & 400k & 14.4 \\
  \quad + DiverseDiT & >458M & 400k & 16.1 \\
  \quad + SRA2 & 458M & 400k & 14.3 \\
  \rowcolor{lightblue}\quad + \ours & 458M & 400k &  \textbf{13.3} \\
  \midrule
  SiT-XL/2 & 675M & 400k & 17.2 \\
  \quad + DiverseDiT & 713M & 400k & 12.4 \\
  \quad + SRA 2 & 675M & 400k & 11.7 \\%
  \rowcolor{lightblue}\quad + \ours & 675M & 400k & \textbf{10.3} \\%
  \midrule
  SiT-XL/2 & 675M & 7M & 8.3 \\
  SiT-XL/2 & 675M & 1M & 12.2 \\
  \quad + SRA2 & 675M & 1M & 8.2 \\%
  \rowcolor{lightblue}\quad + \ours & 675M & 1M & \textbf{7.5} \\%
  \midrule
  SiT-XL/2 & 675M & 2M & 10.1 \\
  \quad + SRA2 & 675M & 2M & 7.0 \\%
  \rowcolor{lightblue}\quad + \ours & 675M & 2M & \textbf{6.6} \\%
  \midrule
  SiT-XL/2 & 675M & 4M & 9.0 \\
  \quad + Disp. & 675M & 4M & 7.7 \\%
  \quad + SRA & 675M & 4M & 7.4 \\%
  \quad + SRA2 & 675M & 4M & 6.6 \\%
  \rowcolor{lightblue}\quad + \ours & 675M & 4M & \textbf{6.2} \\%

  \bottomrule
  \end{tabular}
\end{minipage}
\hfill
\begin{minipage}[t]{0.56\linewidth}
  \vspace{0pt}
  \centering
  \fontsize{8}{9}\selectfont
  \setlength{\tabcolsep}{2.2pt}
  \renewcommand{\arraystretch}{0.95}
  \captionof{table}{Comparison with SoTA methods under CFG.
  }
  \label{tab:rq1-main-comparison}
  \begin{tabular}{lcccccc}
  \toprule
  Method & Epochs & FID$\downarrow$ & sFID$\downarrow$ & IS$\uparrow$ & Pre.$\uparrow$ & Rec.$\uparrow$ \\
  \midrule
  \multicolumn{7}{@{}l}{\textit{Pixel diffusion}} \\
  ADM-U~\citeyearpar{dhariwal2021diffusion} & 400 & 3.94 & 6.14 & 186.7 & 0.82 & 0.52 \\
  VDM++~\citeyearpar{kingma2023understanding} & 560 & 2.40 & - & 225.3 & - & - \\
  Simple diffusion~\citeyearpar{hoogeboom2023simple} & 800 & 2.77 & - & 211.8 & - & - \\
  CDM~\citeyearpar{ho2022classifierfree} & 2160 & 4.88 & - & 158.7 & - & - \\
  \midrule
  \multicolumn{7}{@{}l}{\textit{Latent diffusion, U-Net}} \\
  LDM-4~\citeyearpar{rombach2022ldm} & 200 & 3.60 & - & 247.7 & 0.87 & 0.48 \\
  \midrule
  \multicolumn{7}{@{}l}{\textit{Latent diffusion, Transformer + U-Net hybrid}} \\
  U-ViT-H/2~\citeyearpar{bao2023all} & 240 & 2.29 & 5.68 & 263.9 & 0.82 & 0.57 \\
  DiffiT~\citeyearpar{hatamizadeh2024diffit} & - & 1.73 & - & 276.5 & 0.80 & 0.62 \\
  MDTv2-XL/2~\citeyearpar{gao2023mdtv2} & 1080 & 1.58 & 4.52 & 314.7 & 0.79 & 0.65 \\
  \midrule
  \multicolumn{7}{@{}l}{\textit{Latent diffusion, Transformer}} \\
  MaskDiT~\citeyearpar{zheng2023maskdit} & 1600 & 2.28 & 5.67 & 276.6 & 0.80 & 0.61 \\
  SD-DiT~\citeyearpar{zhu2024sddit} & 480 & 3.23 & - & - & - & - \\
  DiT-XL/2~\citeyearpar{peebles2023dit} & 1400 & 2.27 & 4.60 & 278.2 & 0.83 & 0.57 \\
  \quad + TRADE~\citeyearpar{krause2025tread} & 740 & 1.69 & 4.70 & 305.7 & 0.80 & 0.65 \\
  SiT-XL/2~\citeyearpar{ma2024sit} & 1400 & 2.06 & 4.50 & 270.3 & 0.82 & 0.59 \\
  \multicolumn{7}{@{}l}{\quad \textit{With external representations}} \\
  \quad + REPA~\citeyearpar{yu2024representation} & 800 & 1.42 & 4.70 & 305.7 & 0.80 & 0.65 \\
  \quad + REG~\citeyearpar{wu2025reg} & 800 & 1.36 & 4.25 & 299.4 & 0.77 & 0.66 \\
  \multicolumn{7}{@{}l}{\quad \textit{Without external representations}} \\
  \quad + LayerSync\textsuperscript{*}~\citeyearpar{haghighi2025layersync} & 800 & 1.56 & 4.82 & 287.0 & 0.80 & 0.63 \\
  \quad + SRA~\citeyearpar{jiangrepresentation} & 800 & 1.58 & 4.65 & 305.7 & 0.80 & 0.65 \\
  \quad + SRA2~\citeyearpar{wang2026SRA2} & 800 & 1.52 & 4.63 & 316.2 & 0.82 & 0.62 \\

  \rowcolor{lightblue}\quad + \ours & 800 & \textbf{1.45} & 4.68 & 306.6 & 0.81 & 0.63 \\
  \bottomrule
  \end{tabular}
  \par\vspace{1pt}\scriptsize\raggedright
  \textsuperscript{*}LayerSync includes an additional reverse-process variance head~\citeyearpar{nichol2021improved}; we remove it for fair comparison.
\end{minipage}
\vspace{0mm}
\end{table}

%% file: sections/5_related.tex
\section{Related work}
  \vspace{-1mm}
\textbf{Diffusion transformers for image generation.}
Diffusion~\citep{ho2020denoising,song2020score,rombach2022ldm,podell2023sdxl} and flow-based models~\citep{lipmanflow,liuflow,esser2024scaling} have become dominant paradigms for visual generation.
With transformer backbones, diffusion models have scaled more effectively to larger architectures and stronger visual generation performance.
DiT~\citep{peebles2023dit} first demonstrated that vision transformers provide a strong backbone for diffusion modeling,
while SiT~\citep{ma2024sit} unified diffusion and flow-based training within a scalable stochastic-interpolant transformer framework.
Despite these advances, diffusion transformer training remains costly, motivating representation-level regularization for more efficient and effective optimization.

\vspace{0mm plus 1mm}
\textbf{Representation regularization for diffusion models.}
Representation regularization has shown promise in improving diffusion training.
MaskDiT~\citep{zheng2023maskdit} and SD-DiT~\citep{zhu2024sddit} adopt masked pretraining objectives, while MAEtok~\citep{chen2025masked} combines reconstruction and alignment losses in tokenizer training.
These approaches often require additional decoders or teacher branches~\citep{jiangrepresentation}.
REPA~\citep{yu2024representation} and iREPA~\citep{singh2025whatmatters} instead align diffusion features with external pretrained representations.
Without external encoders, SRA~\citep{jiangrepresentation} and LayerSync~\citep{haghighi2025layersync} use deeper internal features as alignment targets, DUPA~\citep{peng2026dupa} aligns low-frequency conditions across independently noised paths, and LSEP~\citep{yun2025lsep} promotes linear separability without alignment targets.
Dispersive Loss~\citep{wang2025diffuse} and DiverseDiT~\citep{yang2026diversedit} explore negative-only feature dispersion.
However, InfoNCE-based contrastive regularization, which jointly aligns positive pairs and disperses negative pairs, remains underexplored in diffusion transformer training despite its success in discriminative representation learning.

%% file: sections/6_conclusion.tex
\section{Conclusion and limitation}
In this work, we present \ours, a successful attempt to introduce full InfoNCE-based contrastive learning regularization in diffusion transformers training.
We find that vanilla contrastive learning fails because it tends to exploit the noise components in diffusion inputs to enlarge representation distances.
As a result, it learns noise-induced dispersion rather than proper inter-sample relationships. To address this issue, \ours~anchors the maximum negative-pair distance in InfoNCE using pairwise distances among clean representations, thereby suppressing the noise-induced dispersion shortcut.
As a result, \ours~unlocks the benefit of contrastive learning for diffusion transformer training and outperforms existing representation-alignment methods.

\vspace{0mm plus 1mm}
\noindent\textbf{Limitation and possible remedy.}
\ours~requires estimating clean representation anchors during training, introducing a overhead from an additional near-clean forward pass.
This cost is comparable to methods that require extra feature extraction, such as REPA~\citep{yu2024representation}, and does not affect inference time.
A possible remedy for this overhead is to amortize anchor estimation $m_\mathsf{cran}$ with an EMA update when training samples are near the clean endpoint, which we leave for future work.

%% file: sit_cl/appendix.tex
\section{SDE sampler}
For stochastic sampling, we additionally need the score function $\mathbf{s}(\mathbf{x}, t) = \nabla_{\mathbf{x}} \log p_{t}(\mathbf{x})$. The reverse-time SDE can be written as
\begin{equation}
d\mathbf{x}_{t}
= \mathbf{v}(\mathbf{x}_{t}, t)\, dt
- \frac{1}{2} w_{t}\mathbf{s}(\mathbf{x}_{t}, t)\, dt
+ \sqrt{w_{t}}\, d\overline{\mathbf{W}}_{t},
\label{eq:reverse_sde}
\end{equation}
where $w_t$ is the diffusion coefficient and $\overline{\mathbf{W}}_{t}$ is a reverse-time Wiener process~\citep{song2020score,ma2024sit,wu2025reg}. Since $\epsilon$ is Gaussian, the score satisfies
\begin{equation}
\mathbf{s}(\mathbf{x}, t)
= -\sigma_{t}^{-1}\mathbb{E}\left[\epsilon \mid \mathbf{x}_{t} = \mathbf{x}\right]
= \sigma_{t}^{-1}
\frac{
\alpha_{t}\mathbf{v}(\mathbf{x}, t) - \dot{\alpha}_{t}\mathbf{x}
}{
\alpha_{t}\dot{\sigma}_{t} - \dot{\alpha}_{t}\sigma_{t}
}.
\label{eq:score_from_velocity}
\end{equation}
Following the setup of SiT, we use a linear interpolant to construct the noise path, with $\alpha_{t} = 1 - t$ and $\sigma_{t} = t$, where $t \in [0,1]$.

\section{Additional ablations and diagnostics}
\label{apendix:More ablation studies and analysis}

\textbf{Directly probing the noise shortcut.}
We freeze each SiT-B/2 backbone and train a linear probe to reconstruct the injected noise $\epsilon$ from intermediate representations. We report the normalized mean squared error
$\mathrm{NMSE}=\lVert\hat\epsilon-\epsilon\rVert_2^2/\lVert\epsilon\rVert_2^2$; lower NMSE indicates that the input noise is more accurately decoded and therefore more strongly encoded in the representation.

\begin{table}[H]
\centering
\small
\setlength{\tabcolsep}{14pt}
\caption*{NMSE of linear probes that reconstruct input noise from intermediate representations.}
\begin{tabular}{cccc}
\toprule
Noise level $t$ & SiT & InfoNCE & \ours \\
\midrule
0.10 & 0.8950 & 0.8232 & 0.8789 \\
0.40 & 0.3638 & 0.3150 & 0.3417 \\
0.80 & 0.0288 & 0.0277 & 0.028 \\
\bottomrule
\end{tabular}
\end{table}

Noise is linearly decodable from diffusion representations at all three levels. Relative to SiT, InfoNCE consistently lowers NMSE, indicating greater use of the injected noise, whereas \ours~weakens this amplification. These results support the noise-shortcut hypothesis rather than claiming that noise is the only task-irrelevant cue that contrastive learning may exploit.

\textbf{Weakening the negative term does not improve generation quality.}
Intuitively, another alternative for alleviating over-dispersion and weakened alignment is to directly adjust the weight of the negative-pair term.

\begin{figure}[!htbp]
  \centering
  \includegraphics[width=\linewidth]{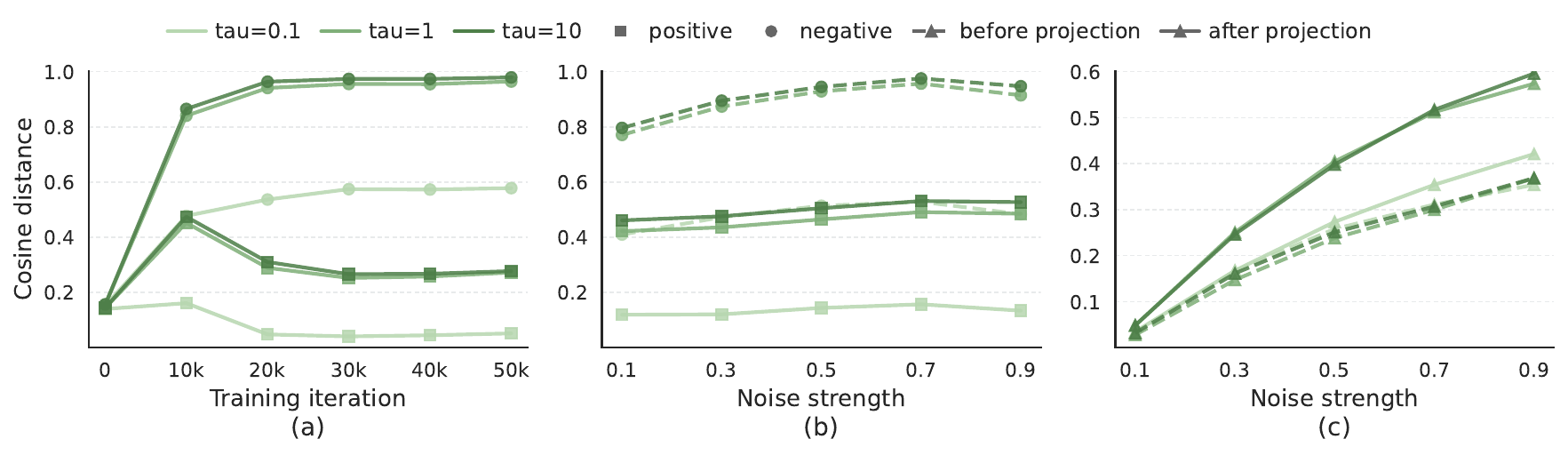}
  \caption{(a) Setting $\tau=0.1$ alleviates over-dispersion and weakened alignment.
(b) Setting $\tau=0.1$ does not change the overall trend that over-dispersion and weakened alignment become more severe as the noise level increases.
(c) For a fixed clean image, using a smaller $\tau$ cannot prevent the contrastive learning head from exploiting noise to increase the distance between representations.}
  \label{fig:temperature_analysis}
\end{figure}

As shown in Figure~\ref{fig:temperature_analysis} (a), using a smaller $\tau$ alleviates over-dispersion and weakened alignment in the contrastive space.
However, it does not lead to a significant performance improvement as shown in Table~\ref{tab:temperature_analysis}. Although using a smaller $\tau$ alleviates over-dispersion and weakened alignment, Figure~\ref{fig:temperature_analysis}(b) shows that, under different values of $\tau$, the representation distance still increases as the noise level grows. In contrast, as shown in Figure~\ref{fig:fig1}(b), when using \ours, the representation distance tends to remain stable, and even decreases with increasing noise when $t>0.5$.
In Figure~\ref{fig:temperature_analysis}(c), the representation distance induced by noise is still amplified by the contrastive projection head, and the degree of amplification further increases as the noise level grows. These observations suggest that reducing $\tau$ does not resolve the underlying representation issue behind over-dispersion and weakened alignment, which explains why simply decreasing $\tau$ brings only limited improvement.

\begin{table}[H]
  \centering
  \small
  \caption{Union of the temperature and anchor ablations. Temperature scaling alone brings only marginal improvements, whereas limiting negative-pair separation with a fixed or adaptive anchor is substantially more effective.}
  \label{tab:fid-comparison-tau}
  \label{tab:temperature_analysis}
  \begin{tabular}{lccc}
  \toprule[1.2pt]
  Model & $\tau$ & Anchor & FID $\downarrow$ \\
  \midrule
  SiT-B/2 & -- & -- & 33.0 \\
  \midrule
  + Pos. reg. & -- & -- & 28.7 \\
  + InfoNCE & 10.0 & -- & 28.3 \\
  + InfoNCE & 5.0 & -- & 28.6 \\
  + InfoNCE & 1.0 & -- & 28.5 \\
  + InfoNCE & 0.5 & -- & 28.9 \\
  + InfoNCE & 0.1 & -- & 28.3 \\
  + InfoNCE & 1.0 & $\cos 60^\circ$ & 27.0 \\
  + \ours & 1.0 & $m_\mathsf{cran}$ & \textbf{26.3} \\
  \bottomrule
  \end{tabular}
\end{table}

\textbf{The effectiveness of clean image representation distances as the hinge.}
Next, we study how diffusion representation distances at different noise levels affect the ability of \ours~to correctly learn inter-sample relationships.
Specifically, in addition to clean features, we use the inter-sample representation distances at $t=0.5$ and $t=0.9$ as the maximum distance in \ours~optimization.
As shown in Table~\ref{tab:mclean_ablation}, when the noise weight in the samples becomes too large, the corresponding representation distances no longer provide reliable guidance for \ours~training.

\begin{table}[H]
\caption{FID-50K of different feature regularization methods on SiT-B/2 trained on ImageNet 256$\times$256 for 400K iterations.}
\label{tab:mclean_ablation}
\centering
\begin{tabular}{llll}
\toprule[1.2pt]
Model             & iter. & FID $\downarrow$ & IS $\uparrow$ \\
\midrule
SiT-B/2           & 400k & 33.0 & 43.7 \\
\midrule

SiT-B/2 + \ours      &   &   &  \\
\quad \textbf{t=0.1}      & 400k & \textbf{26.3} & \textbf{56.1} \\
\quad t=0.5      & 400k & 26.7 & 55.6 \\
\quad t=0.9      & 400k & 28.2 & 50.8 \\
\bottomrule
\end{tabular}
\end{table}

\textbf{Student--teacher layer pairing.}
We next vary the student and teacher layers on SiT-B/2. Changing the student layer or moving the teacher to a shallower layer clearly degrades performance, whereas moving the teacher from layer 8 to the deeper layer 10 has only a modest effect.

\begin{table}[H]
\centering
\small
\caption{Ablation of the student and teacher layers on SiT-B/2.}
\label{tab:layer-pairing}
\begin{tabular}{ccccccc}
\toprule
Student & Teacher & FID$\downarrow$ & sFID$\downarrow$ & IS$\uparrow$ & Pre.$\uparrow$ & Rec.$\uparrow$ \\
\midrule
2 & 8 & 44.8 & 11.5 & 33.5 & 0.49 & 0.62 \\
3 & 8 & \textbf{42.9} & \textbf{11.4} & \textbf{36.0} & \textbf{0.50} & \textbf{0.62} \\
4 & 8 & 44.2 & \textbf{11.4} & 33.7 & \textbf{0.50} & \textbf{0.62} \\
3 & 6 & 51.2 & 13.4 & 30.3 & 0.45 & 0.61 \\
3 & 10 & \textbf{42.9} & 11.7 & 35.4 & \textbf{0.50} & 0.61 \\
\bottomrule
\end{tabular}
\end{table}

\input{sit_cl/appendix_generalization}
\input{sit_cl/appendix_efficiency_complementarity}
\input{sit_cl/appendix_clean_geometry_estimation}

\section{Experimental setup}

\label{appendix:Experimental setups}
\textbf{Hyperparameter settings.}
Our default hyperparameter settings are summarized in Table~\ref{tab:hyperparameter-settings}.

\begin{table}[H]
\centering
\caption{Hyperparameter settings.}
\label{tab:hyperparameter-settings}
\small
\begin{tabular}{@{}lccc@{}}
\toprule
\textbf{Backbone} & \textbf{SiT-B} & \textbf{SiT-L} & \textbf{SiT-XL} \\
\midrule
\multicolumn{4}{@{}l}{\textbf{Architecture}} \\
Input & $32 \times 32 \times 4$ & $32 \times 32 \times 4$ & $32 \times 32 \times 4$ \\
Patch size & 2 & 2 & 2 \\
Layers & 12 & 24 & 28 \\
Hidden dim. & 768 & 1,024 & 1,152 \\
Num. heads & 12 & 16 & 16 \\
\midrule
\multicolumn{4}{@{}l}{\textbf{\ours~settings}} \\
\multicolumn{4}{@{}l}{\textit{Contrastive between internal representations}} \\
Contrastive blocks & 8 $\leftrightarrow$ 3 & 16 $\leftrightarrow$ 6 & 20 $\leftrightarrow$ 8 \\
Contrastive time interval & [0, 0.2) & [0, 0.2) & [0, 0.2) \\
$\lambda$ & 1.0 & 1.0 & 1.0 \\
\multicolumn{4}{@{}l}{\textit{Contrastive between internal representations and DINO representations}} \\
\midrule
\multicolumn{4}{@{}l}{\textbf{Optimization}} \\
Batch size & 256 & 256 & 256 \\
Optimizer & AdamW & AdamW & AdamW \\
lr & 0.0001 & 0.0001 & 0.0001 \\
$(\beta_1, \beta_2)$ & (0.9, 0.999) & (0.9, 0.999) & (0.9, 0.999) \\
\midrule
\multicolumn{4}{@{}l}{\textbf{Interpolants}} \\
$\alpha_t$ & $1-t$ & $1-t$ & $1-t$ \\
$\sigma_t$ & $t$ & $t$ & $t$ \\
$w_t$ & $\sigma_t$ & $\sigma_t$ & $\sigma_t$ \\
Training objective & v-prediction & v-prediction & v-prediction \\
Sampler & Euler--Maruyama & Euler--Maruyama & Euler--Maruyama \\
Sampling steps & 250 & 250 & 250 \\
Guidance Scale &-&-&1.8 \\
\bottomrule
\end{tabular}
\end{table}

\textbf{Computational resources.}
All training experiments are conducted on 8 NVIDIA H200 GPUs, with a global batch size of 256 and 32 samples per GPU.

\section{Metrics}
\label{appendix:metrics}
We evaluate image generation quality from complementary perspectives using standard quantitative metrics.

\textbf{Fr\'echet Inception Distance (FID)}~\citep{heusel2017gans} measures the distributional similarity between real and generated images in the feature space of an Inception-v3 network.
Lower FID indicates that generated samples are closer to the real data distribution.

\textbf{Structural FID (sFID)}~\citep{nash2021sparse} extends FID by computing the distance on spatial feature maps, making it more sensitive to structural and spatial coherence.

\textbf{Inception Score (IS)}~\citep{salimans2016improved} evaluates both image quality and class-conditional diversity by measuring the discrepancy between confident per-image class predictions and the marginal class distribution over generated samples.
Higher IS is better.

\textbf{Precision and recall}~\citep{kynkaanniemi2019improved} separate sample fidelity from distribution coverage.
Precision reflects how realistic generated samples are relative to real images, while recall measures how well the generated samples cover the target data manifold.

To investigate how \ours~affects learned representations, in addition to linear probing accuracy, we follow iREPA~\citep{singh2025whatmatters} and use spatial structure metrics to measure patch-level organization.

\textbf{Spatial self-similarity metrics.}
These metrics analyze whether intermediate representations preserve spatial structure.
Given patch features $\mathbf{X}=\{\mathbf{x}_{t}\}_{t=1}^{T}$ arranged on an $H\times W$ lattice, we define the cosine self-similarity kernel as
$K_{\mathbf{X}}(t,t')=\langle \mathbf{x}_{t},\mathbf{x}_{t'}\rangle/(\|\mathbf{x}_{t}\|_{2}\|\mathbf{x}_{t'}\|_{2})$.
The Manhattan distance between two patch locations is denoted as $d(t,t')$.

\textbf{Local vs.\ Distant Similarity (LDS)} measures the contrast between local and distant patch similarities:
\begin{equation}
\mathrm{LDS}(\mathbf{X})
=
\mathbb{E}\!\left[K_{\mathbf{X}}(t,t') \mid d(t,t') < r_{\mathrm{near}}\right]
-
\mathbb{E}\!\left[K_{\mathbf{X}}(t,t') \mid d(t,t') \ge r_{\mathrm{far}}\right].
\end{equation}
A larger LDS indicates that nearby patches are more similar than far-away patches, suggesting stronger spatial organization.

\textbf{Correlation Decay Slope (CDS)} quantifies how fast patch similarity decays with spatial distance.
We first compute the spatial correlogram $g_{\mathbf{X}}(\delta)=\mathbb{E}[K_{\mathbf{X}}(t,t') \mid d(t,t')=\delta]$ for $\delta\in\{1,\ldots,\Delta\}$.
After fitting a least-squares line $g_{\mathbf{X}}(\delta)\approx \alpha+\beta\delta$, CDS is defined as
\begin{equation}
\mathrm{CDS}(\mathbf{X})=-\hat{\beta}.
\end{equation}
Higher CDS corresponds to a sharper decay of similarity over distance and therefore stronger spatial structure.

\textbf{RMS Spatial Contrast (RMSC)} measures the spatial variation of normalized patch features.
Let $\hat{\mathbf{x}}_{t}=\mathbf{x}_{t}/\|\mathbf{x}_{t}\|_{2}$ and $\bar{\mathbf{x}}=\frac{1}{T}\sum_{t=1}^{T}\hat{\mathbf{x}}_{t}$.
RMSC is defined as
\begin{equation}
\mathrm{RMSC}(\mathbf{X})
=
\sqrt{\frac{1}{T}\sum_{t=1}^{T}\|\hat{\mathbf{x}}_{t}-\bar{\mathbf{x}}\|_{2}^{2}}.
\end{equation}
Higher RMSC indicates stronger spatial diversity across patch tokens, while lower RMSC suggests that the representation becomes spatially uniform.

\section{Uncurated Samples}
\label{appendix:Samples}
We show uncurated samples generated by SiT-XL with \ours~on ImageNet $256 \times 256$, using classifier-free guidance with $w=4.0$.

\begin{figure}[!htbp]
  \centering
  \includegraphics[width=\linewidth]{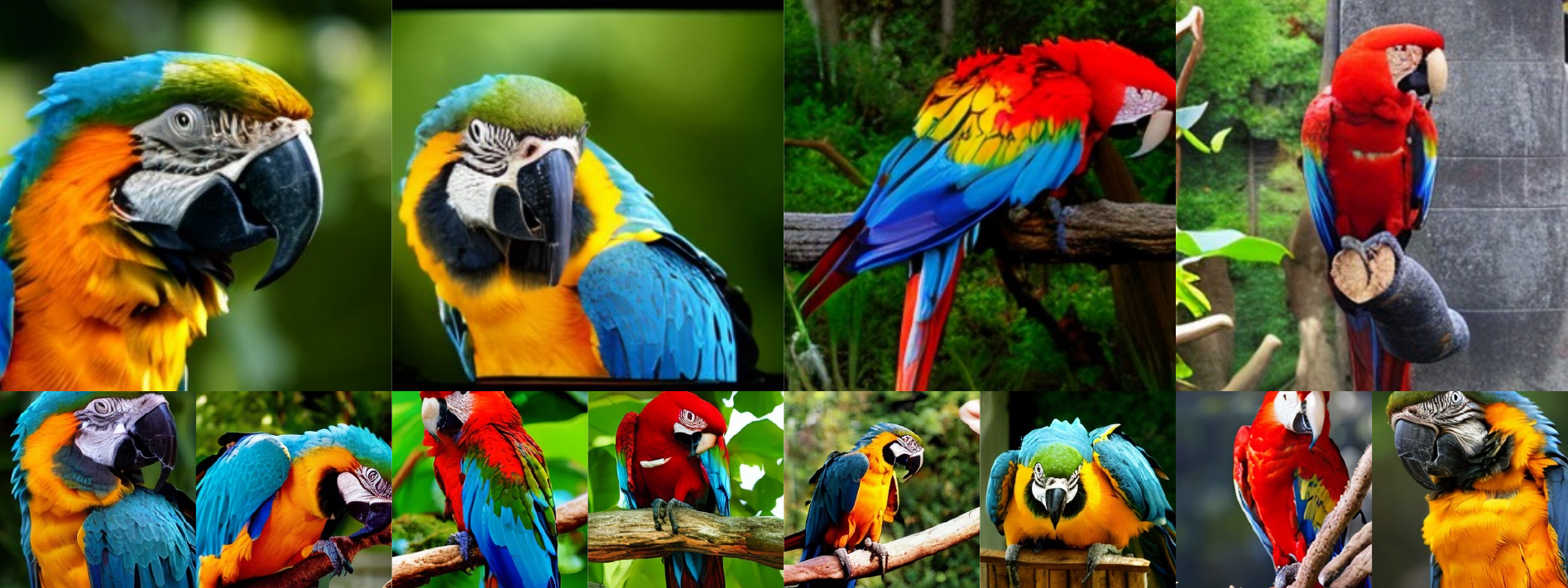}
  \caption{Generated images from SiT-XL/2 + \ours~conditioned on the class label ``macaw'' (88).}
\end{figure}

\begin{figure}[!htbp]
  \centering
  \includegraphics[width=\linewidth]{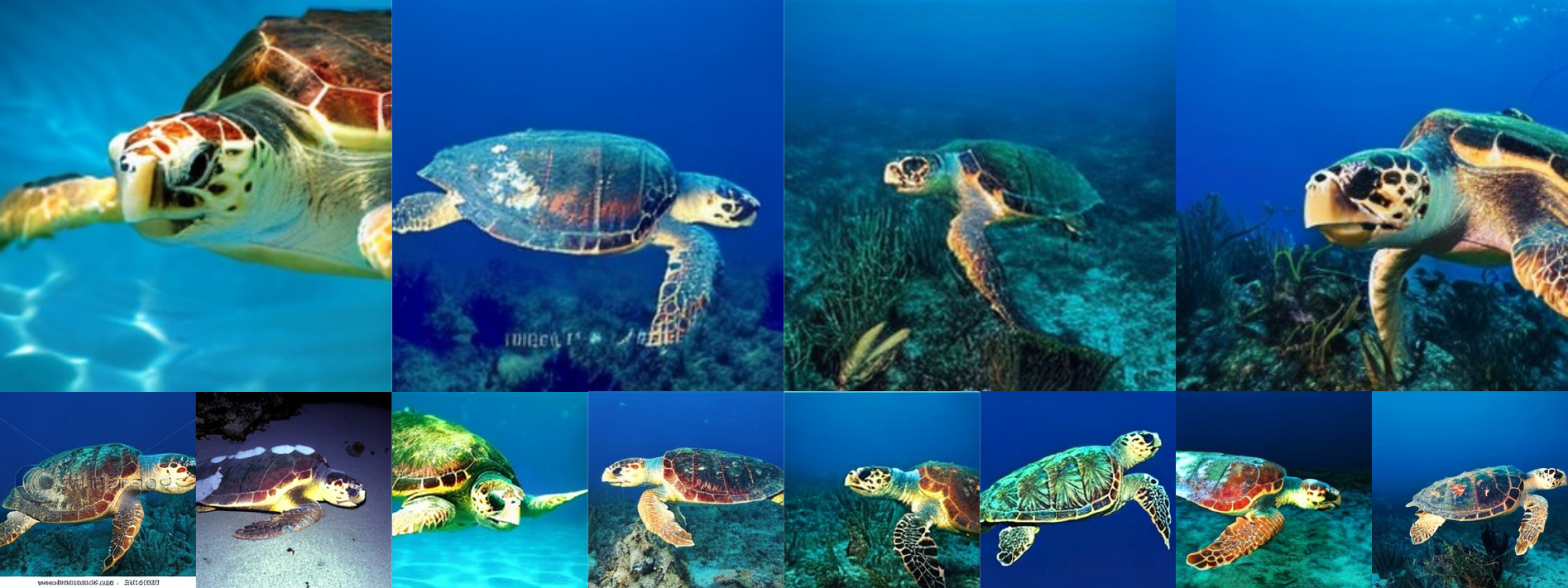}
  \caption{Generated images from SiT-XL/2 + \ours~conditioned on the class label ``loggerhead'' (33).}
\end{figure}

\begin{figure}[!htbp]
  \centering
  \includegraphics[width=\linewidth]{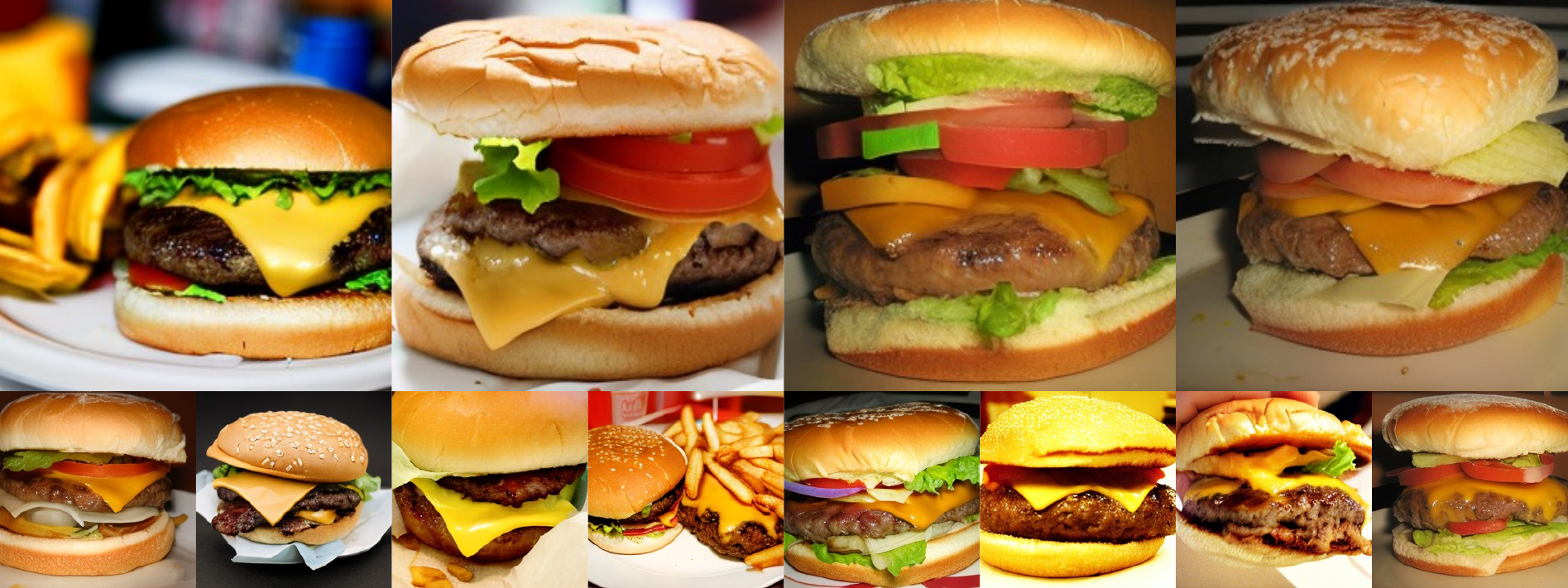}
  \caption{Generated images from SiT-XL/2 + \ours~conditioned on the class label ``cheeseburger'' (933).}
\end{figure}

\begin{figure}[!htbp]
  \centering
  \includegraphics[width=\linewidth]{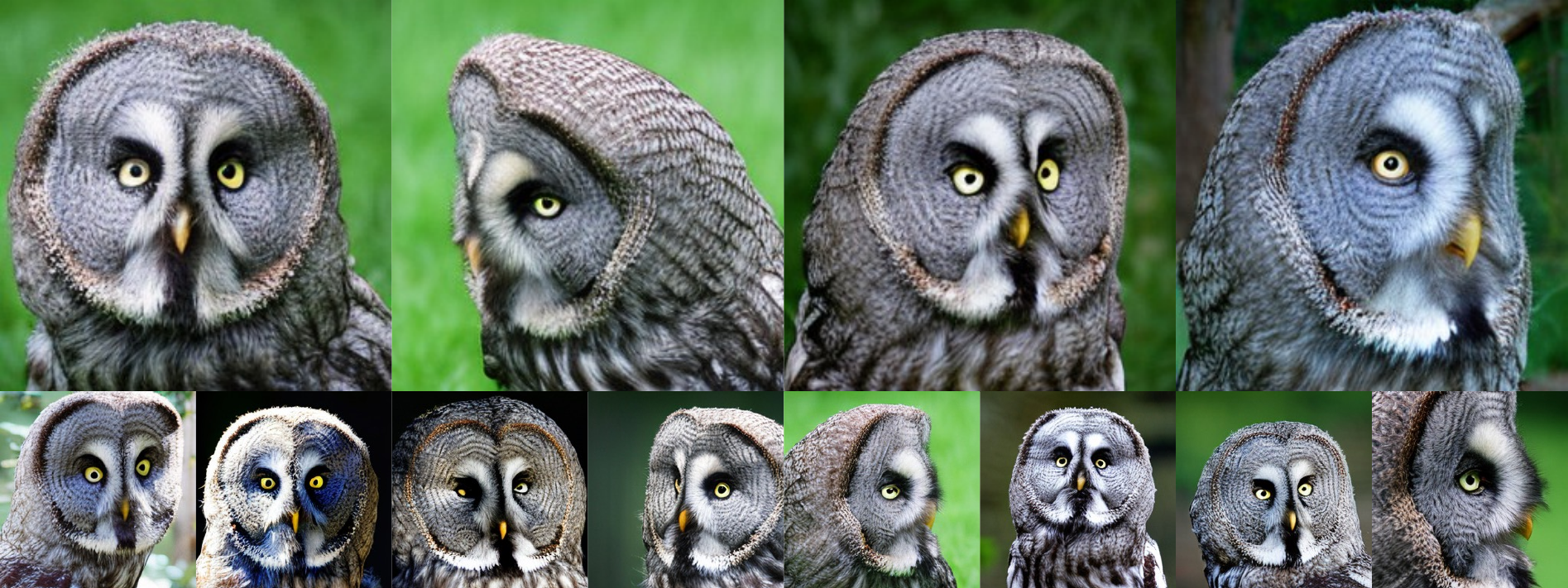}
  \caption{Generated images from SiT-XL/2 + \ours~conditioned on the class label ``great grey owl'' (24).}
\end{figure}

\begin{figure}[!htbp]
  \centering
  \includegraphics[width=\linewidth]{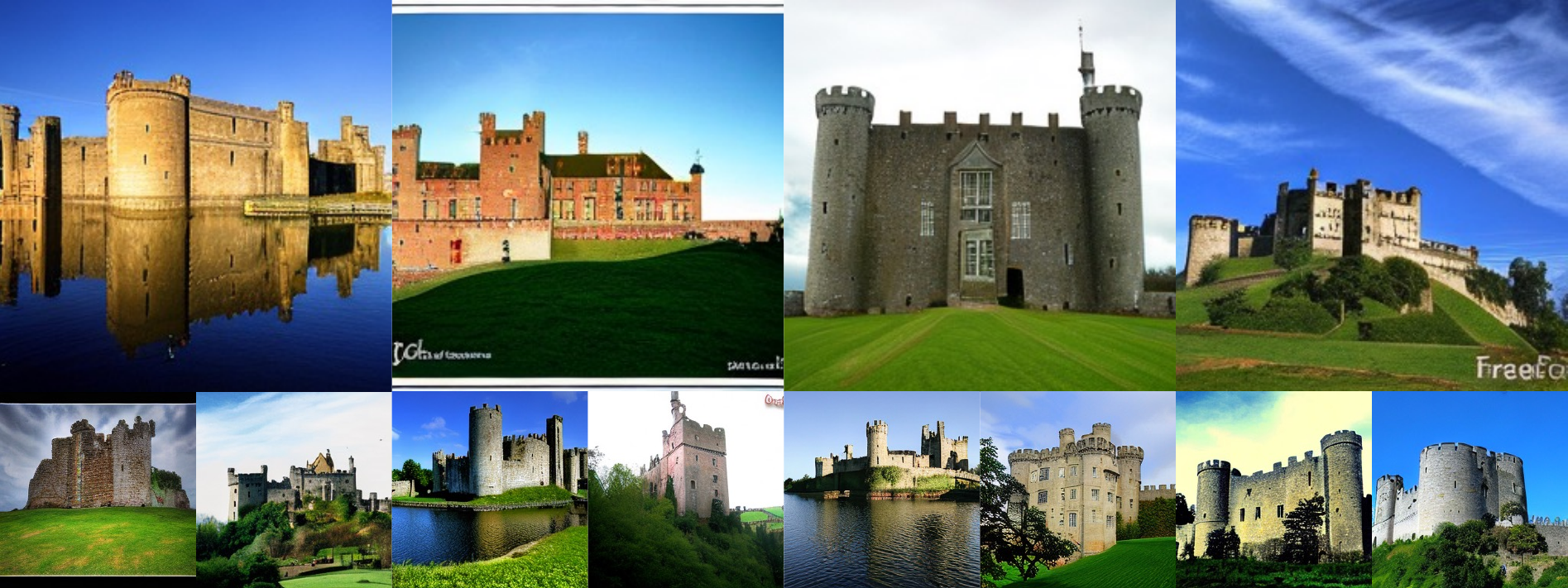}
  \caption{Generated images from SiT-XL/2 + \ours~conditioned on the class label ``castle'' (483).}
\end{figure}

\begin{figure}[!htbp]
  \centering
  \includegraphics[width=\linewidth]{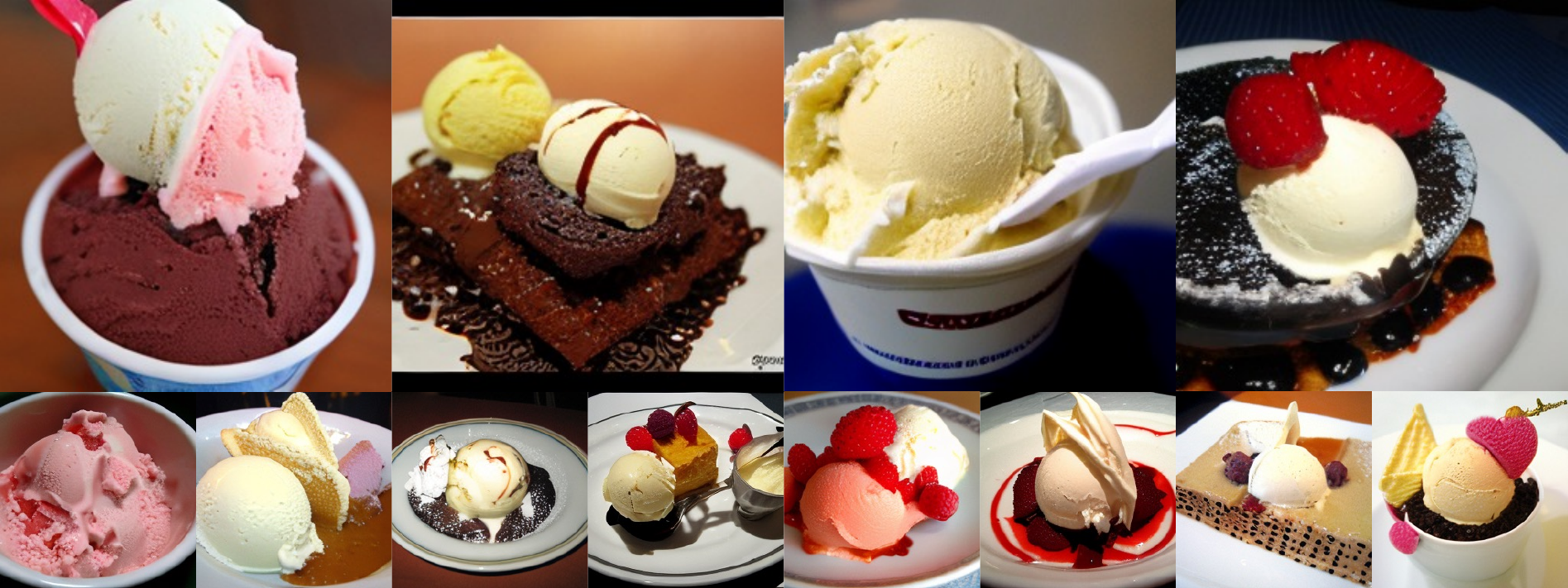}
  \caption{Generated images from SiT-XL/2 + \ours~conditioned on the class label ``ice cream'' (928).}
\end{figure}

\begin{figure}[!htbp]
  \centering
  \includegraphics[width=\linewidth]{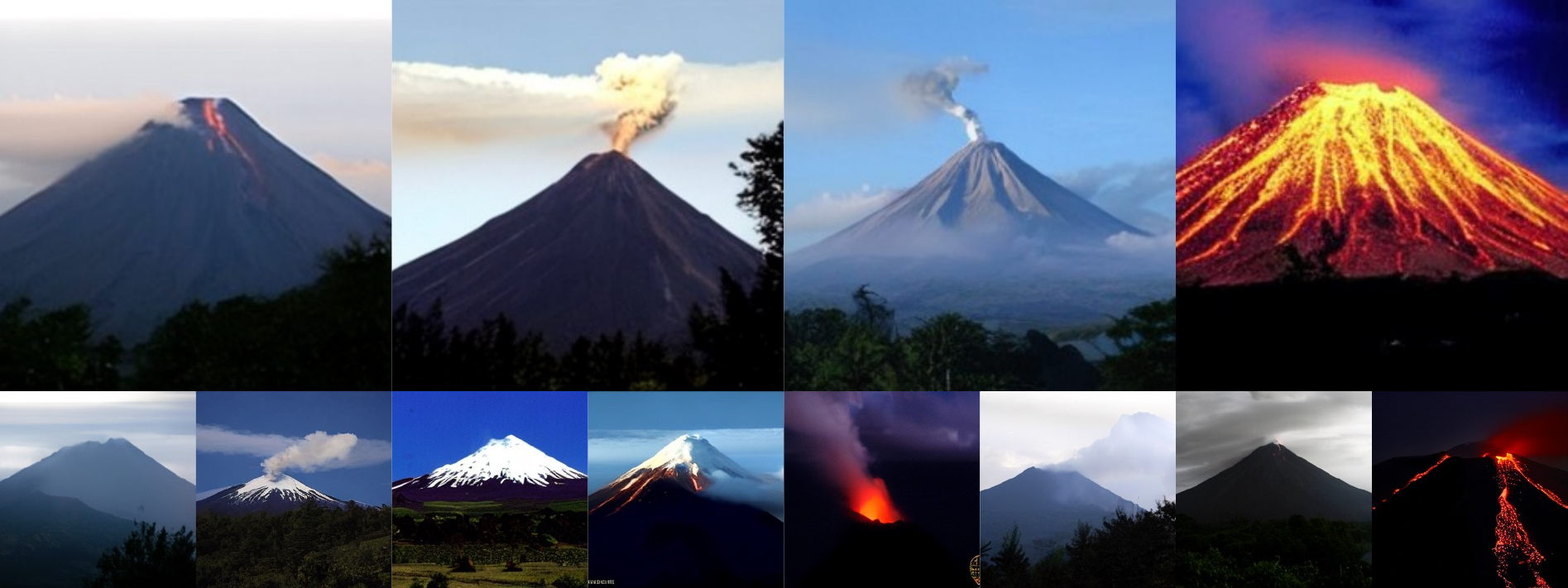}
  \caption{Generated images from SiT-XL/2 + \ours~conditioned on the class label ``volcano'' (980).}
\end{figure}

\begin{figure}[!htbp]
  \centering
  \includegraphics[width=\linewidth]{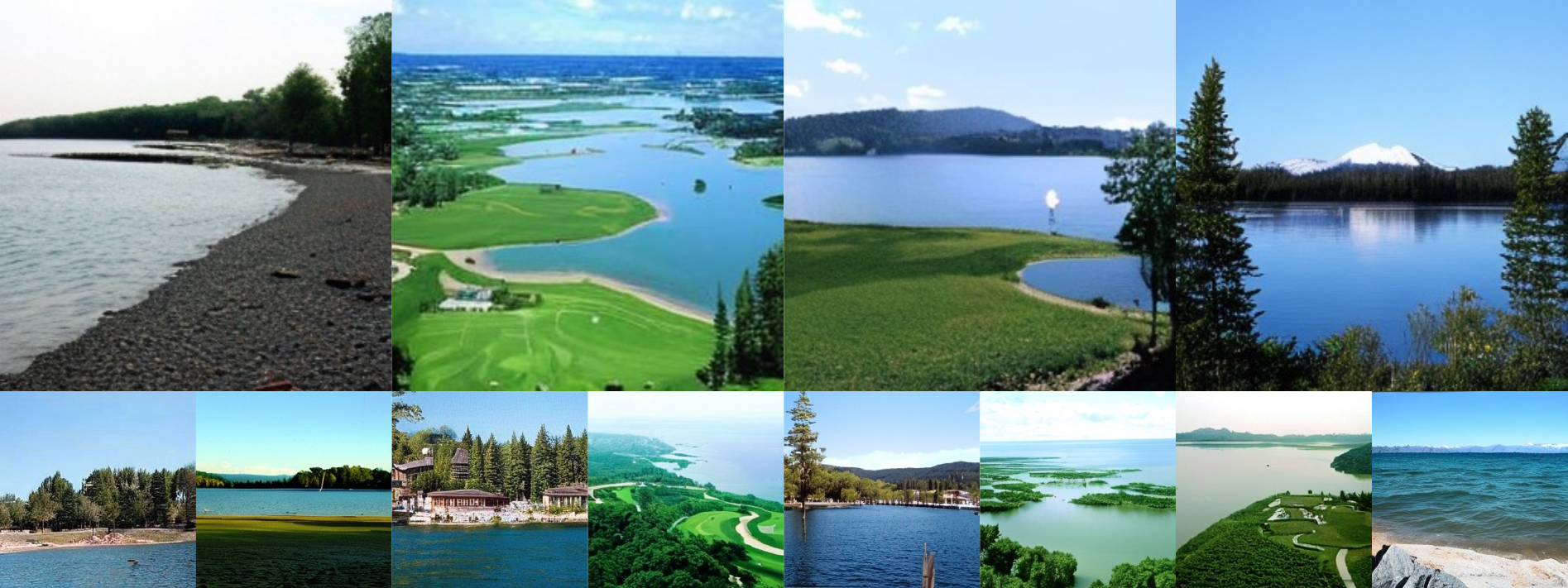}
  \caption{Generated images from SiT-XL/2 + \ours~conditioned on the class label ``lakeside'' (975).}
\end{figure}

\begin{figure}[!htbp]
  \centering
  \includegraphics[width=\linewidth]{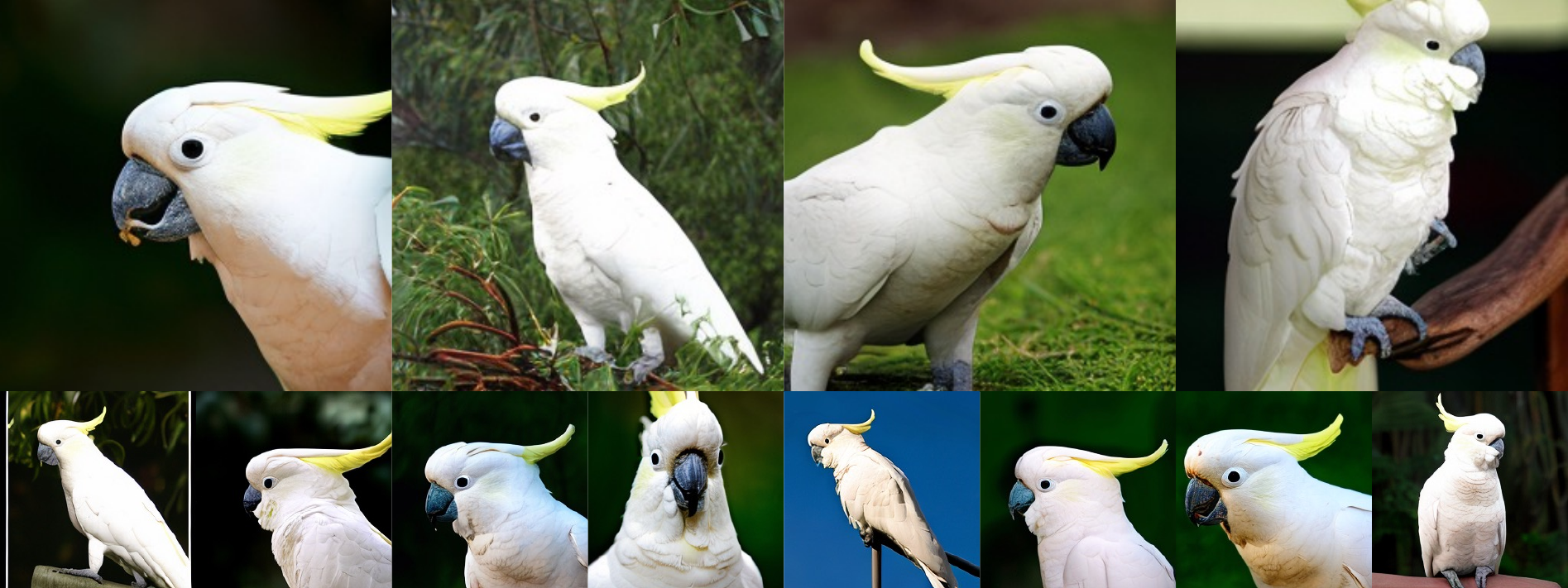}
  \caption{Generated images from SiT-XL/2 + \ours~conditioned on the class label ``sulphur-crested cockatoo'' (89).}
\end{figure}

\begin{figure}[!htbp]
  \centering
  \includegraphics[width=\linewidth]{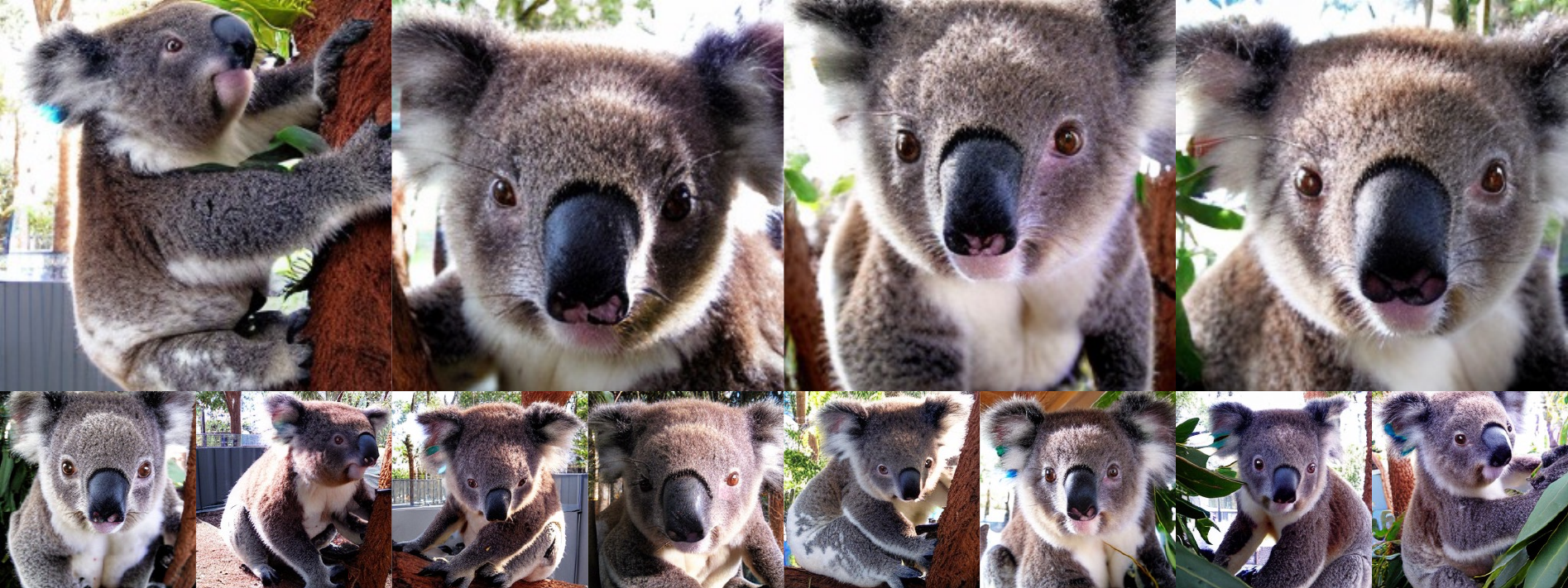}
  \caption{Generated images from SiT-XL/2 + \ours~conditioned on the class label ``koala'' (105).}
\end{figure}

\begin{figure}[!htbp]
  \centering
  \includegraphics[width=\linewidth]{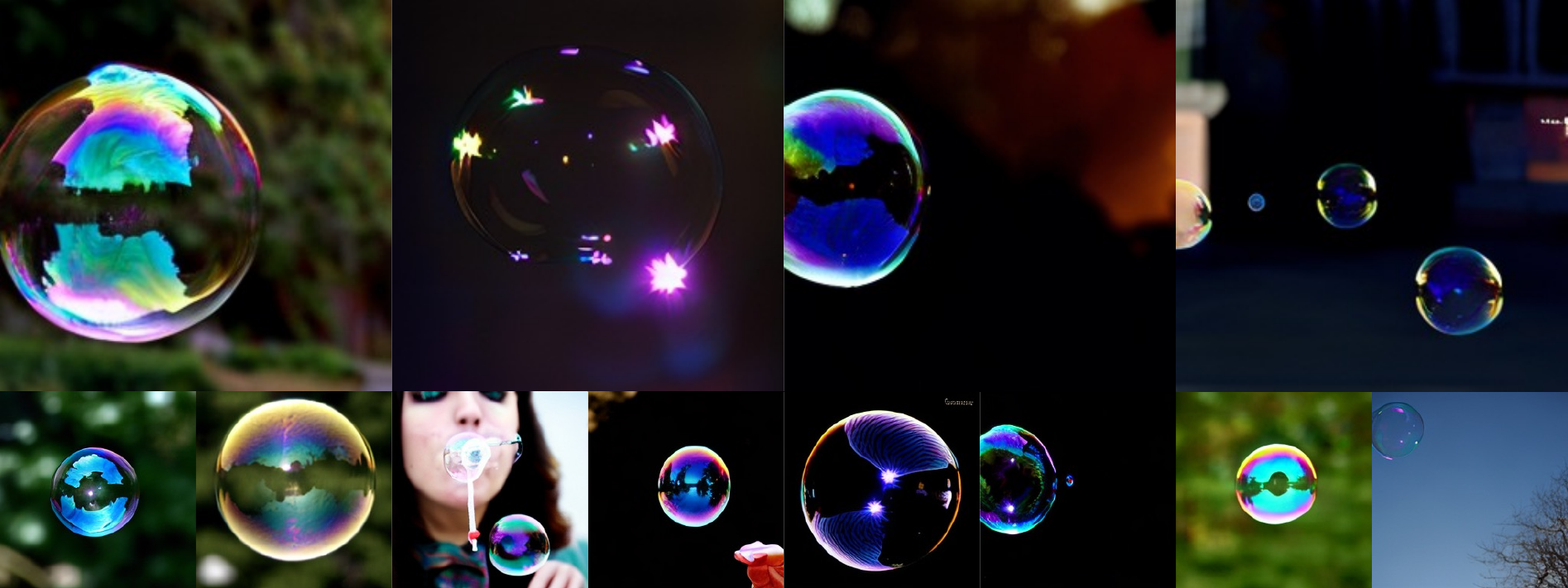}
  \caption{Generated images from SiT-XL/2 + \ours~conditioned on the class label ``bubble'' (971).}
\end{figure}

\begin{figure}[!htbp]
  \centering
  \includegraphics[width=\linewidth]{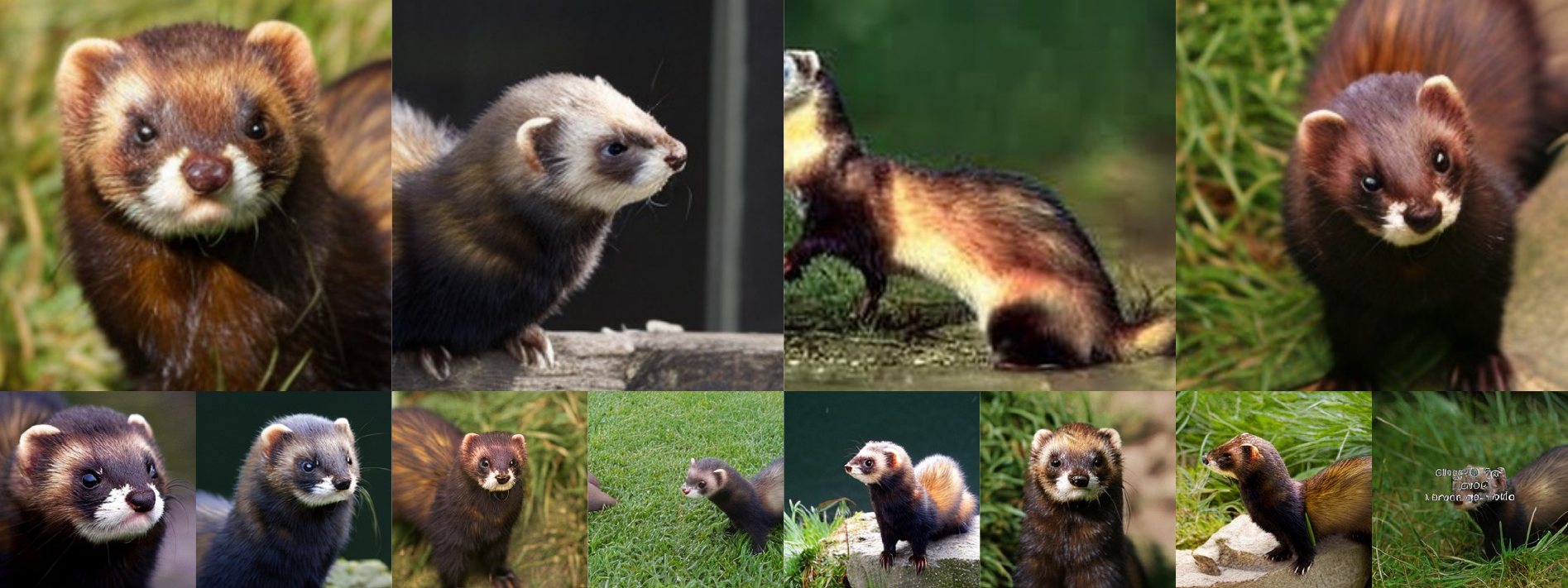}
  \caption{Generated images from SiT-XL/2 + \ours~conditioned on the class label ``polecat'' (358).}
\end{figure}

\begin{figure}[!htbp]
  \centering
  \includegraphics[width=\linewidth]{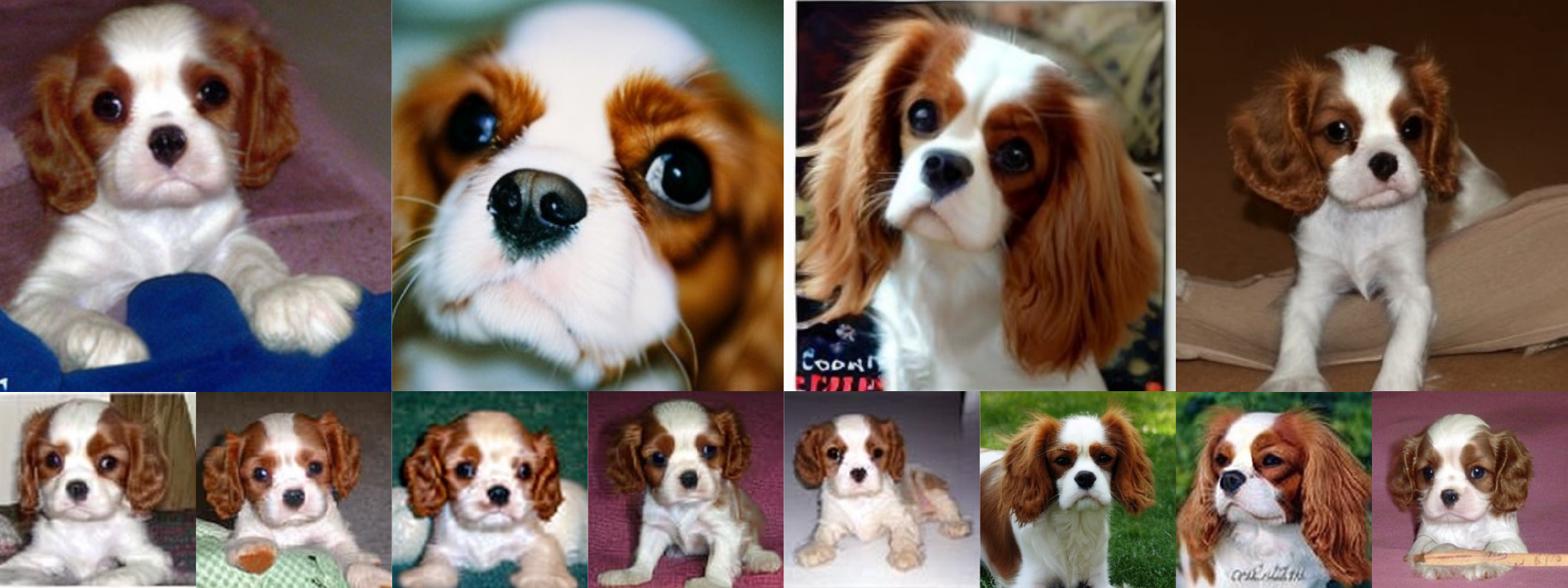}
  \caption{Generated images from SiT-XL/2 + \ours~conditioned on the class label ``Blenheim spaniel'' (156).}
\end{figure}

\begin{figure}[!htbp]
  \centering
  \includegraphics[width=\linewidth]{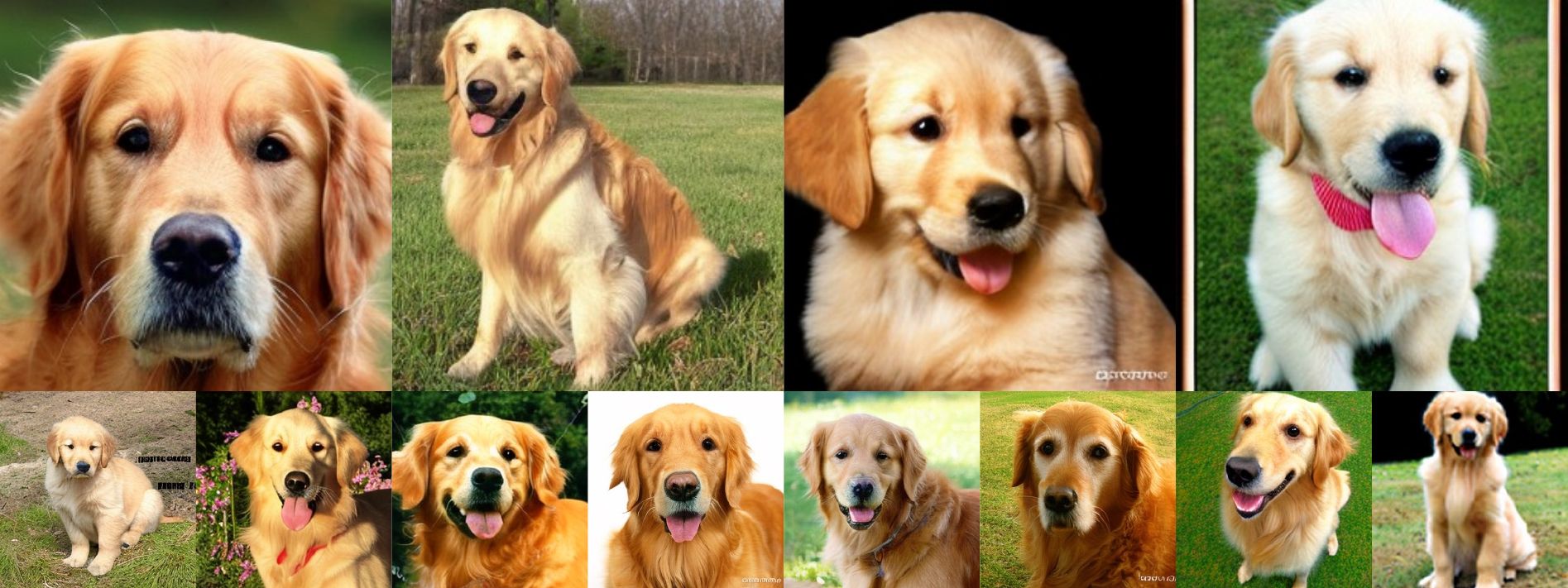}
  \caption{Generated images from SiT-XL/2 + \ours~conditioned on the class label ``golden retriever'' (207).}
\end{figure}

\begin{figure}[!htbp]
  \centering
  \includegraphics[width=\linewidth]{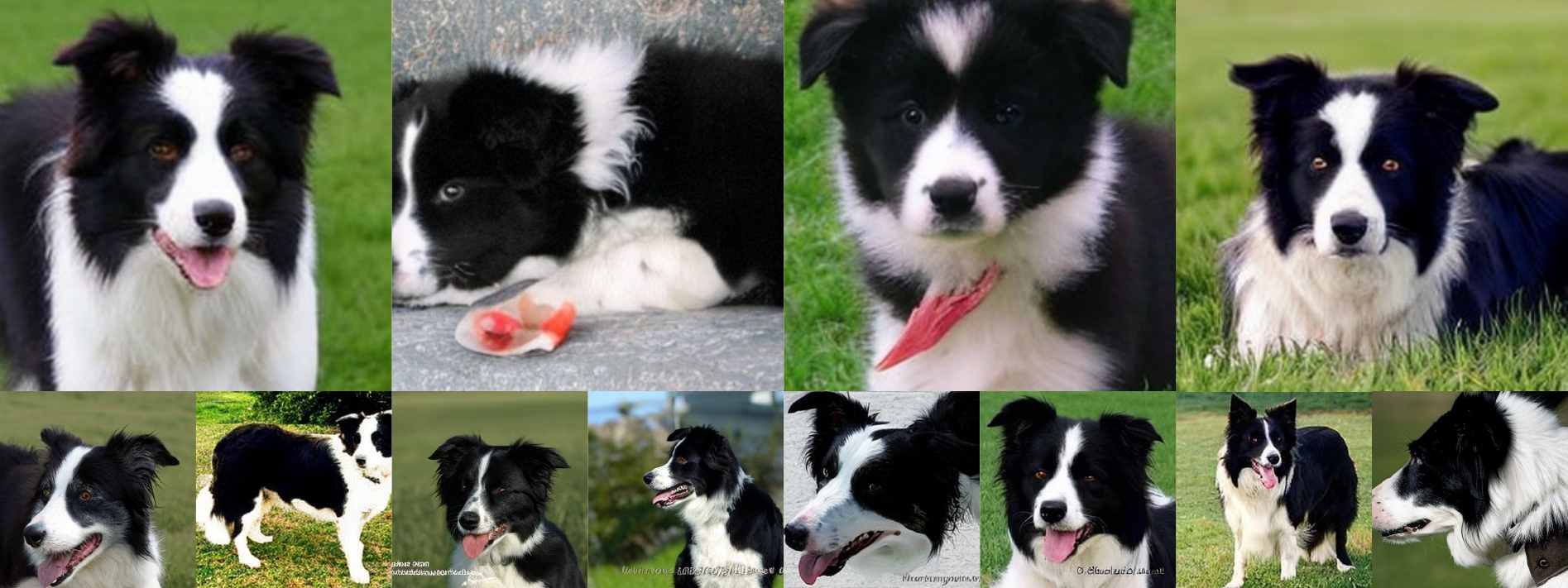}
  \caption{Generated images from SiT-XL/2 + \ours~conditioned on the class label ``Border collie'' (232).}
\end{figure}

\begin{figure}[!htbp]
  \centering
  \includegraphics[width=\linewidth]{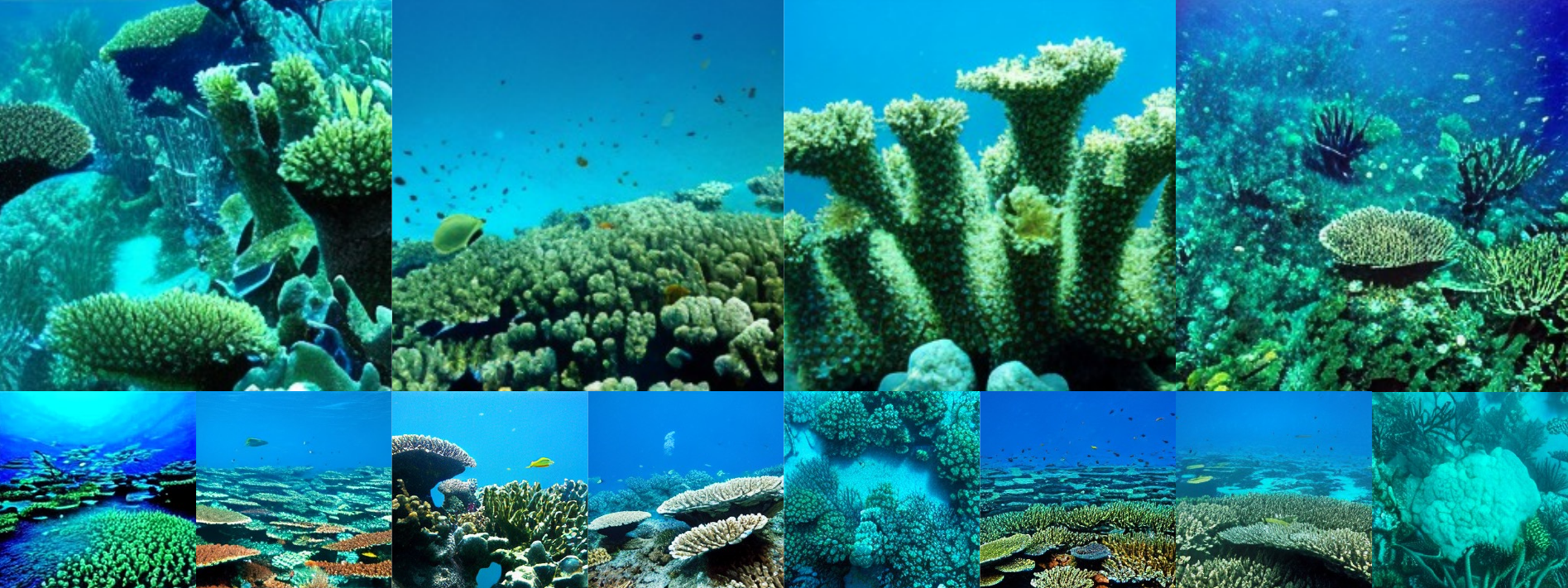}
  \caption{Generated images from SiT-XL/2 + \ours~conditioned on the class label ``coral reef'' (973).}
\end{figure}

\begin{figure}[!htbp]
  \centering
  \includegraphics[width=\linewidth]{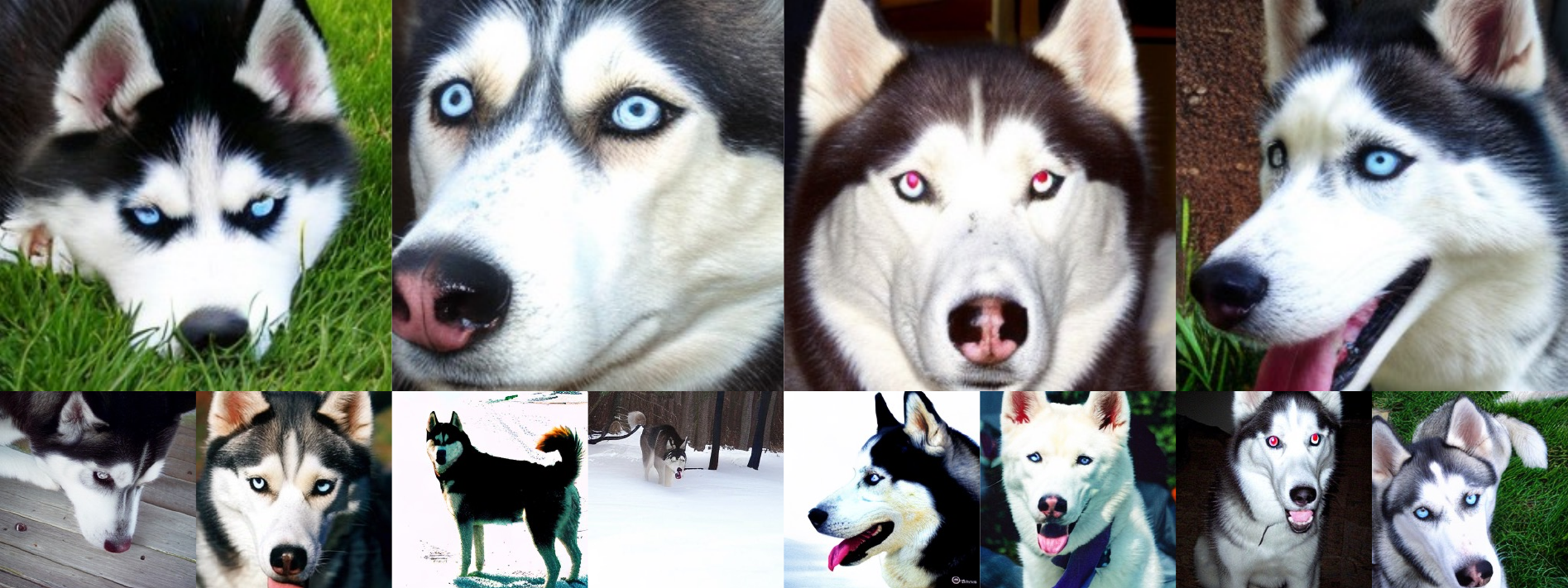}
  \caption{Generated images from SiT-XL/2 + \ours~conditioned on the class label ``Siberian husky'' (250).}
\end{figure}

\begin{figure}[!htbp]
  \centering
  \includegraphics[width=\linewidth]{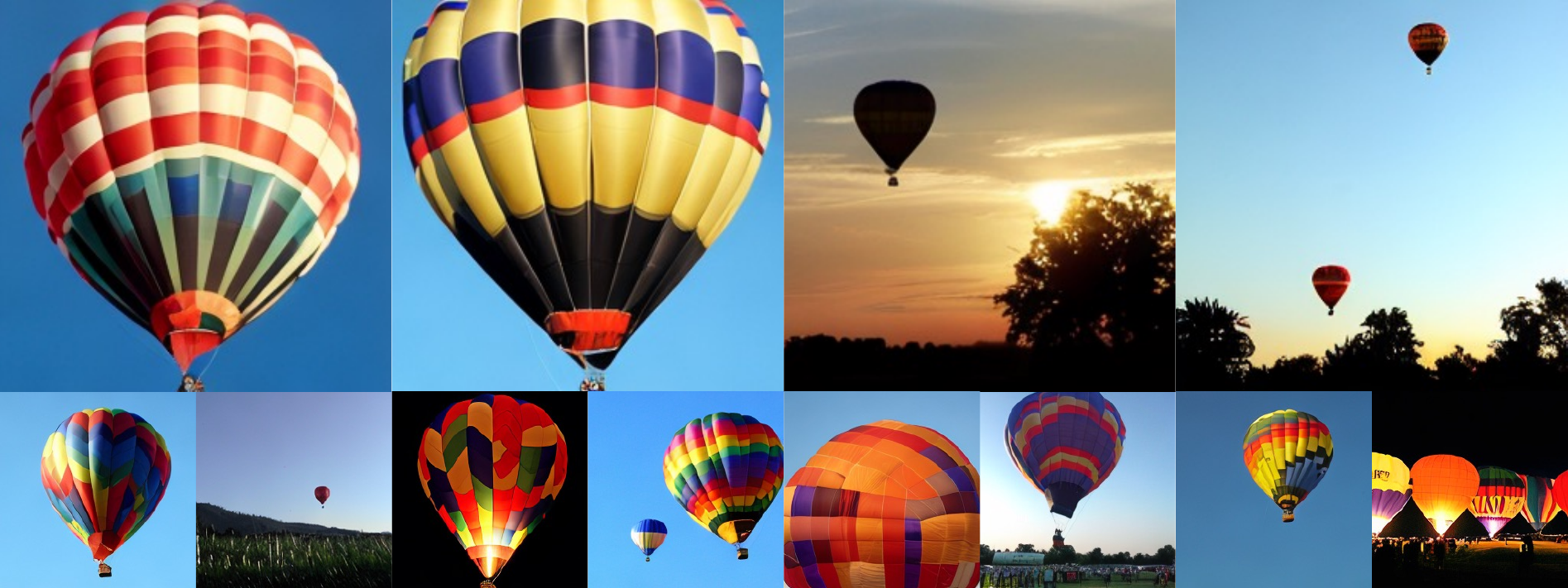}
  \caption{Generated images from SiT-XL/2 + \ours~conditioned on the class label ``balloon'' (417).}
\end{figure}

\begin{figure}[!htbp]
  \centering
  \includegraphics[width=\linewidth]{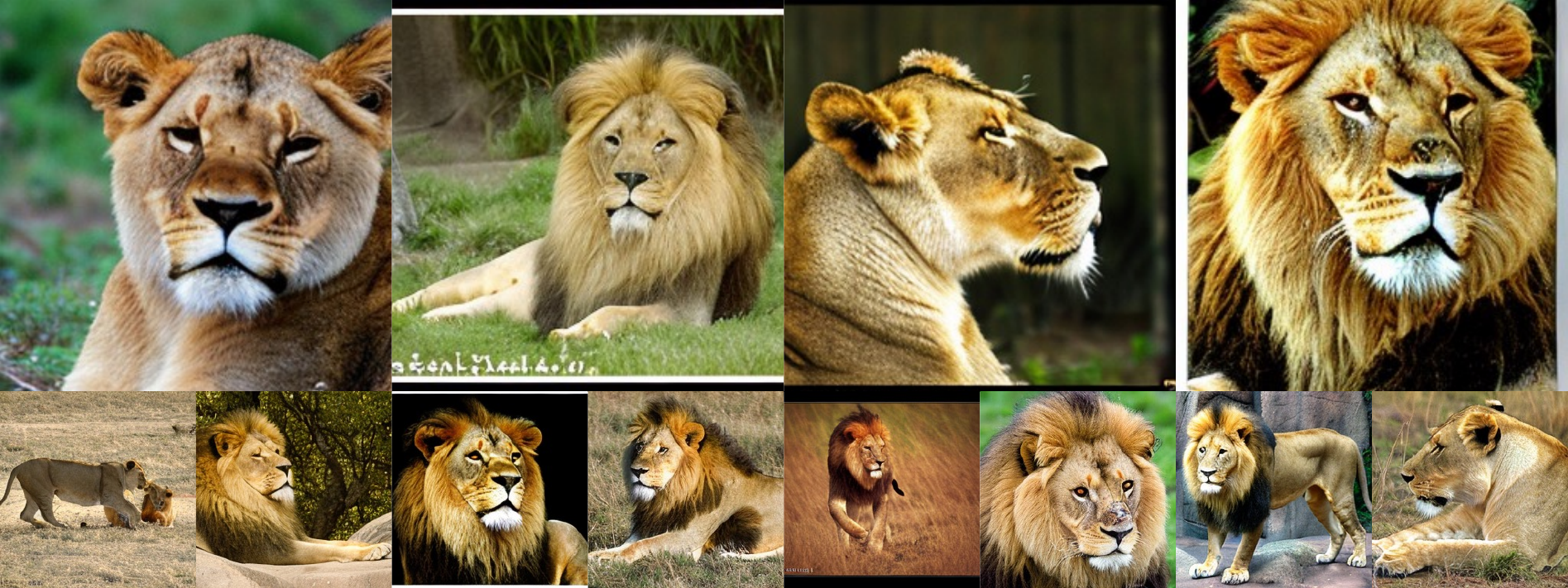}
  \caption{Generated images from SiT-XL/2 + \ours~conditioned on the class label ``lion'' (291).}
\end{figure}

\begin{figure}[!htbp]
  \centering
  \includegraphics[width=\linewidth]{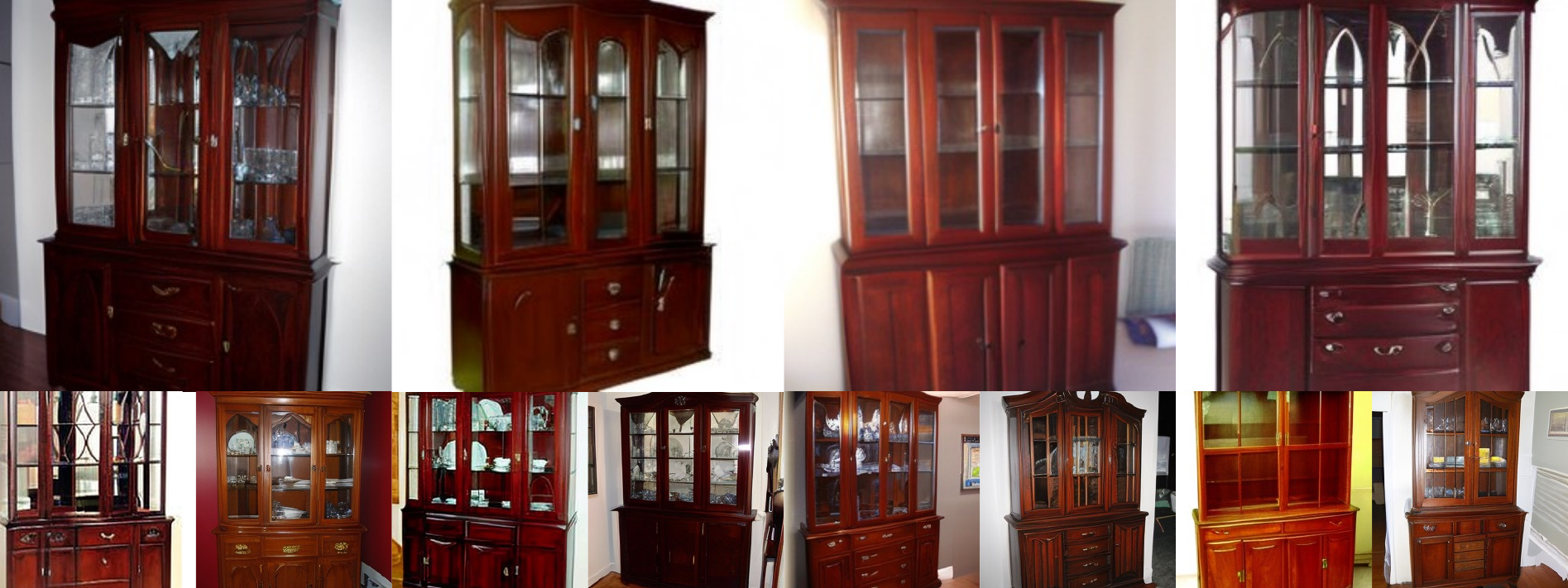}
  \caption{Generated images from SiT-XL/2 + \ours~conditioned on the class label ``china cabinet'' (495).}
\end{figure}

%% file: sit_cl/appendix_generalization.tex
\section{Generalization beyond ImageNet-256}
\label{appendix:generalization}

\subsection{Low-category and structurally homogeneous datasets}
To test whether \ours~depends on the broad semantic diversity of ImageNet-1K, we evaluate it on FFHQ, a structurally homogeneous single-category face dataset. For a CIFAR-10-scale class-conditional setting, the native $32\times32$ resolution of CIFAR-10 is usually studied with U-Net diffusion models and is poorly matched to our focus on diffusion transformers. We therefore construct WebVision-10-50K from WebVision 1.0~\citep{li2017webvision}: we select ten ImageNet-style classes and sample 5,000 images per class, yielding the same number of classes and training images as CIFAR-10 while retaining a $256\times256$ setting.

\begin{table}[H]
\centering
\small
\setlength{\tabcolsep}{3.5pt}
\caption{Results on datasets with substantially lower semantic diversity than ImageNet-1K. All methods use SiT-B/2 and 100K training steps.}
\label{tab:low-category-generalization}
\begin{tabular}{llccccc}
\toprule
Dataset & Method & FID$\downarrow$ & sFID$\downarrow$ & IS$\uparrow$ & Pre.$\uparrow$ & Rec.$\uparrow$ \\
\midrule
FFHQ & SiT & 6.73 & 7.55 & 4.4 & 0.71 & 0.47 \\
FFHQ & REPA & 6.49 & 7.63 & 4.5 & 0.71 & 0.48 \\
FFHQ & \ours & \textbf{5.59} & \textbf{7.02} & \textbf{4.6} & \textbf{0.71} & \textbf{0.49} \\
\midrule
WebVision-10-50K & SiT & 27.6 & 9.84 & 15.2 & 0.45 & 0.47 \\
WebVision-10-50K & \ours & \textbf{20.5} & \textbf{9.45} & \textbf{15.8} & \textbf{0.48} & \textbf{0.47} \\
\bottomrule
\end{tabular}
\end{table}

On FFHQ, \ours~reduces FID by 16.9\% relative to SiT and provides a larger gain than out-of-domain DINO-based alignment. On WebVision-10-50K, it reduces FID by 25.7\% and improves all reported metrics. These results show that the benefit of \ours~does not require a large number of semantic categories.

\subsection{Higher-resolution generation}
We train SiT-B/2 from scratch on ImageNet $512\times512$. Relative to the $256\times256$ setting, the latent patch sequence grows from $16\times16=256$ to $32\times32=1{,}024$ tokens.

\begin{table}[H]
\centering
\small
\caption{ImageNet $512\times512$ results with SiT-B/2.}
\label{tab:imagenet512}
\begin{tabular}{lcccccc}
\toprule
Method & Steps & FID$\downarrow$ & sFID$\downarrow$ & IS$\uparrow$ & Pre.$\uparrow$ & Rec.$\uparrow$ \\
\midrule
SiT & 200K & 56.7 & 8.11 & 26.8 & 0.53 & \textbf{0.62} \\
\ours & 200K & \textbf{45.5} & \textbf{6.98} & \textbf{32.7} & \textbf{0.61} & \textbf{0.62} \\
\midrule
SiT & 400K & 43.3 & 7.41 & 36.8 & 0.61 & \textbf{0.64} \\
\ours & 400K & \textbf{32.9} & \textbf{6.57} & \textbf{47.4} & \textbf{0.67} & 0.63 \\
\bottomrule
\end{tabular}
\end{table}

At 400K steps, \ours~reduces FID from 43.3 to 32.9 and increases IS from 36.8 to 47.4, showing that its benefit persists with a fourfold longer token sequence.

\subsection{Generalization to \texorpdfstring{$\epsilon$}{epsilon}-prediction}
Our main experiments use the SiT velocity objective. To test whether the method depends on that formulation, we apply \ours~to DiT-B/2~\citep{peebles2023dit} with its original $\epsilon$-prediction objective.

\begin{table}[H]
\centering
\small
\caption{Generalization from SiT velocity prediction to DiT $\epsilon$-prediction on ImageNet $256\times256$.}
\label{tab:dit-generalization}
\begin{tabular}{lcccccc}
\toprule
Method & Steps & FID$\downarrow$ & sFID$\downarrow$ & IS$\uparrow$ & Pre.$\uparrow$ & Rec.$\uparrow$ \\
\midrule
DiT-B/2 & 400K & 43.5 & 8.71 & 32.7 & 0.49 & 0.63 \\
DiT-B/2 + \ours & 400K & \textbf{38.6} & \textbf{8.24} & \textbf{37.6} & \textbf{0.51} & \textbf{0.64} \\
\bottomrule
\end{tabular}
\end{table}

The consistent improvement across all five metrics indicates that \ours~is not specific to velocity prediction.

\subsection{Human motion generation}
We further apply \ours~to the Motion Diffusion Model (MDM)~\citep{tevet2023mdm} on HumanML3D~\citep{guo2022humanml3d}, where a domain-matched large-scale pretrained representation encoder is unavailable. For an out-of-domain REPA baseline, we render skeletal motion sequences as RGB skeleton frames and extract DINO representations from those renderings.

\begin{table}[H]
\centering
\small
\caption{Text-to-motion generation on HumanML3D.}
\label{tab:motion-generalization}
\begin{tabular}{lcc}
\toprule
Method & FID$\downarrow$ & R-Precision$\uparrow$ \\
\midrule
MDM & $0.5320\pm0.0464$ & $0.6106\pm0.0044$ \\
MDM + REPA & $0.5400\pm0.0391$ & $0.6063\pm0.0061$ \\
MDM + \ours & $\mathbf{0.3918\pm0.0347}$ & $\mathbf{0.6694\pm0.0075}$ \\
\bottomrule
\end{tabular}
\end{table}

The visual REPA target does not improve MDM in this out-of-domain setting, whereas \ours~substantially improves both distributional quality and text--motion correspondence using only internal diffusion representations.

%% file: sit_cl/appendix_efficiency_complementarity.tex
\section{Efficiency, sampling robustness, and complementarity}
\label{appendix:efficiency-complementarity}

\subsection{Computational overhead across model scales}
We profile SiT, SRA, REPA, and \ours~under matched hardware, global batch size, precision, software, and data-loading settings. Table~\ref{tab:efficiency-profile} reports average per-step time, computation, and peak memory.

\begin{table}[H]
\centering
\small
\setlength{\tabcolsep}{4.5pt}
\caption{Training efficiency across backbone scales.}
\label{tab:efficiency-profile}
\begin{tabular}{clcccc}
\toprule
Backbone & Method & Time/step & Ratio & GFLOPs/sample & Peak memory \\
\midrule
B/2 & SiT & 36.92 ms & 1.000$\times$ & 139.2 & 6.29 GiB \\
B/2 & SRA & 41.19 ms & 1.116$\times$ & 173.5 & 6.41 GiB \\
B/2 & \ours & 41.32 ms & 1.119$\times$ & 173.5 & 6.45 GiB \\
B/2 & REPA & 52.39 ms & 1.419$\times$ & 194.4 & 6.92 GiB \\
\midrule
L/2 & SiT & 99.01 ms & 1.000$\times$ & 487.4 & 18.66 GiB \\
L/2 & SRA & 108.56 ms & 1.096$\times$ & 601.4 & 18.83 GiB \\
L/2 & \ours & 109.50 ms & 1.106$\times$ & 601.4 & 18.84 GiB \\
L/2 & REPA & 115.86 ms & 1.170$\times$ & 545.8 & 19.29 GiB \\
\midrule
XL/2 & SiT & 140.63 ms & 1.000$\times$ & 715.9 & 26.10 GiB \\
XL/2 & SRA & 155.23 ms & 1.104$\times$ & 893.5 & 26.30 GiB \\
XL/2 & \ours & 156.11 ms & 1.110$\times$ & 893.6 & 26.32 GiB \\
XL/2 & REPA & 156.59 ms & 1.114$\times$ & 774.7 & 26.73 GiB \\
\bottomrule
\end{tabular}
\end{table}

The relative per-step overhead of \ours~over SiT remains stable across scales: 11.9\% on B/2, 10.6\% on L/2, and 11.0\% on XL/2. On XL/2, \ours~takes 0.3\% less time per step than REPA while using less peak memory.

\subsection{Matched wall-clock budgets}
Using the measured per-step times in Table~\ref{tab:efficiency-profile}, 400K \ours~steps match 448K SiT steps, while 400K REPA steps match 507K \ours~steps. Each adjacent pair in Table~\ref{tab:matched-time} therefore has the same wall-clock budget.

\begin{table}[H]
\centering
\small
\setlength{\tabcolsep}{3.5pt}
\caption{Generation performance under matched wall-clock budgets on SiT-B/2.}
\label{tab:matched-time}
\begin{tabular}{llcccccc}
\toprule
Budget & Method & Steps & FID$\downarrow$ & sFID$\downarrow$ & IS$\uparrow$ & Pre.$\uparrow$ & Rec.$\uparrow$ \\
\midrule
SiT & SiT & 448K & 33.7 & 6.59 & 44.4 & 0.54 & \textbf{0.64} \\
SiT & \ours & 400K & \textbf{26.3} & \textbf{6.16} & \textbf{56.2} & \textbf{0.58} & 0.63 \\
\midrule
REPA & REPA & 400K & \textbf{24.4} & 6.40 & \textbf{59.9} & \textbf{0.59} & \textbf{0.65} \\
REPA & \ours & 507K & 24.5 & \textbf{6.12} & 59.6 & \textbf{0.59} & 0.64 \\
\bottomrule
\end{tabular}
\end{table}

Under matched time, \ours~clearly outperforms SiT and remains comparable to REPA.

\subsection{Fixed sampling protocol}
Guidance-interval tuning introduces an additional degree of freedom. We therefore compare all methods at CFG scale 1.35, applying guidance over the full sampling trajectory without interval scheduling. All results use the same 250-step Euler--Maruyama SDE sampler and 50K-sample evaluation protocol.

\begin{table}[H]
\centering
\small
\caption{Comparison under a shared CFG scale and sampling protocol.}
\label{tab:fixed-cfg}
\begin{tabular}{lccccc}
\toprule
Method & FID$\downarrow$ & sFID$\downarrow$ & IS$\uparrow$ & Pre.$\uparrow$ & Rec.$\uparrow$ \\
\midrule
REPA & \textbf{1.80} & 4.50 & 284.0 & 0.81 & \textbf{0.61} \\
SRA & 1.89 & 4.47 & 278.7 & 0.81 & \textbf{0.61} \\
\ours & 1.83 & \textbf{4.35} & \textbf{286.6} & \textbf{0.83} & 0.59 \\
\bottomrule
\end{tabular}
\end{table}

Under this fixed protocol, \ours~is comparable to REPA and outperforms SRA overall, with a remaining recall trade-off.

\subsection{CFG and guidance-interval sweeps}
For completeness, Table~\ref{tab:cfg-sweeps} reports the expanded sweeps used to select the balanced operating point in the main comparison. The main \ours~result uses CFG scale 1.8 and interval $[0,0.675]$.

\begin{table}[H]
\centering
\scriptsize
\setlength{\tabcolsep}{3pt}
\caption{CFG-scale and guidance-interval sweeps. The REPA entries are its published sweep; SRA is swept around its official setting.}
\label{tab:cfg-sweeps}
\begin{tabular}{lccccccc}
\toprule
Method & Interval & CFG & FID$\downarrow$ & sFID$\downarrow$ & IS$\uparrow$ & Pre.$\uparrow$ & Rec.$\uparrow$ \\
\midrule
\ours & $[0,0.65]$ & 1.70 & 1.52 & 4.76 & 277.4 & 0.80 & 0.62 \\
\ours & $[0,0.65]$ & 1.80 & 1.49 & 4.78 & 289.6 & 0.81 & 0.62 \\
\ours & $[0,0.65]$ & 1.90 & 1.54 & 4.83 & 299.5 & 0.81 & 0.61 \\
\ours & $[0,0.675]$ & 1.70 & \textbf{1.44} & 4.75 & 295.5 & 0.81 & 0.64 \\
\ours & $[0,0.675]$ & 1.80 & 1.45 & 4.68 & 306.6 & 0.81 & 0.63 \\
\ours & $[0,0.675]$ & 1.90 & 1.64 & 4.71 & 323.9 & 0.82 & 0.61 \\
\ours & $[0,0.70]$ & 1.80 & 1.63 & 4.56 & 316.0 & 0.82 & 0.61 \\
\midrule
REPA & $[0,0.60]$ & 2.00 & 1.56 & 5.11 & 290.7 & 0.78 & 0.66 \\
REPA & $[0,0.65]$ & 2.00 & 1.44 & 4.88 & 308.8 & 0.79 & 0.65 \\
REPA & $[0,0.70]$ & 1.80 & \textbf{1.42} & 4.70 & 305.7 & 0.80 & 0.64 \\
REPA & $[0,0.70]$ & 1.90 & 1.45 & 4.68 & 317.6 & 0.80 & 0.64 \\
REPA & $[0,0.70]$ & 2.00 & 1.48 & 4.67 & 324.0 & 0.82 & 0.62 \\
REPA & $[0,0.75]$ & 2.00 & 1.78 & 4.50 & 346.2 & 0.82 & 0.62 \\
REPA & $[0,0.80]$ & 2.00 & 2.23 & 4.40 & 360.9 & 0.84 & 0.60 \\
\midrule
SRA & $[0,0.675]$ & 1.80 & 1.57 & 4.76 & 301.5 & 0.79 & 0.64 \\
SRA & $[0,0.70]$ & 1.70 & \textbf{1.57} & 4.70 & 294.9 & 0.79 & 0.64 \\
SRA & $[0,0.70]$ & 1.80 & 1.58 & 4.65 & 309.4 & 0.80 & 0.63 \\
SRA & $[0,0.70]$ & 1.90 & 1.69 & 4.61 & 325.9 & 0.81 & 0.63 \\
SRA & $[0,0.725]$ & 1.80 & 1.67 & 4.56 & 320.3 & 0.81 & 0.63 \\
\bottomrule
\end{tabular}
\end{table}

\subsection{Complementarity with REPA}
REPA aligns each noisy diffusion feature with a corresponding external DINOv2 target learned for discriminative visual understanding. In contrast, the internal teacher used by \ours~is optimized through the generation objective and retains generation-relevant layout, texture, instance detail, and denoising-trajectory information. The optimization signals also differ: REPA provides a sample-wise pull toward a fixed external target, whereas \ours~adds relational gradients determined by other samples and the adaptive clean anchor. These gradients are generally not collinear, making the two objectives complementary rather than redundant.

\begin{table}[H]
\centering
\small
\caption{Adding \ours~to REPA on SiT-B/2.}
\label{tab:repa-complementarity}
\begin{tabular}{clccccc}
\toprule
Steps & Method & FID$\downarrow$ & sFID$\downarrow$ & IS$\uparrow$ & Pre.$\uparrow$ & Rec.$\uparrow$ \\
\midrule
100K & REPA & 49.5 & 7.00 & 27.5 & 0.46 & 0.59 \\
100K & REPA + \ours & \textbf{42.4} & \textbf{6.64} & \textbf{32.5} & \textbf{0.50} & \textbf{0.60} \\
\midrule
200K & REPA & 33.2 & \textbf{6.68} & 43.7 & 0.54 & \textbf{0.63} \\
200K & REPA + \ours & \textbf{28.5} & 6.87 & \textbf{51.4} & \textbf{0.57} & \textbf{0.63} \\
\midrule
400K & REPA & 24.4 & \textbf{6.40} & 59.9 & 0.59 & \textbf{0.65} \\
400K & REPA + \ours & \textbf{21.0} & 6.46 & \textbf{68.6} & \textbf{0.60} & \textbf{0.65} \\
\bottomrule
\end{tabular}
\end{table}

Adding \ours~improves FID, IS, and precision at every budget. At 400K steps, it reduces FID from 24.4 to 21.0 while preserving recall.

%% file: sit_cl/appendix_clean_geometry_estimation.tex
\section{Effectiveness of Clean Inter-Sample Geometry Estimation}
\label{appendix:clean-geometry-estimation}

Our default implementation estimates $m_{\mathsf{cran}}$ from the low-noise samples already present in the current batch, thereby avoiding an additional near-clean forward pass. Table~\ref{tab:estimated-mcran-fid} shows that this efficient estimate preserves generation quality: its FID of 26.3 is within 0.1 of the full $m_{\mathsf{cran}}$ computation.

\begin{table}[H]
\centering
\small
\setlength{\tabcolsep}{5pt}
\caption{FID comparison on ImageNet $256\times256$ using SiT-B/2, 400K training steps, and no CFG. The estimated $m_{\mathsf{cran}}$ variant is our default.}
\label{tab:estimated-mcran-fid}
\begin{tabular}{lccc}
\toprule
Model & \#Params & Iter. & FID$\downarrow$ \\
\midrule
SiT-B/2 & 130M & 400k & 33.0 \\
\quad + SRA & 130M & 400k & 29.1 \\
\quad + SRA \& Disp. & 130M & 400k & 28.6 \\
\quad + DiverseDiT & $>$130M & 400k & 28.1 \\
\quad + SRA2 & 130M & 400k & 28.9 \\
\quad + \ours, full $m_{\mathsf{cran}}$ & 130M & 400k & \textbf{26.2} \\
\rowcolor{lightblue}\quad + \ours, estimated $m_{\mathsf{cran}}$ & 130M & 400k & 26.3 \\
\bottomrule
\end{tabular}
\end{table}

\begin{figure}[H]
  \centering
  \includegraphics[width=0.86\linewidth]{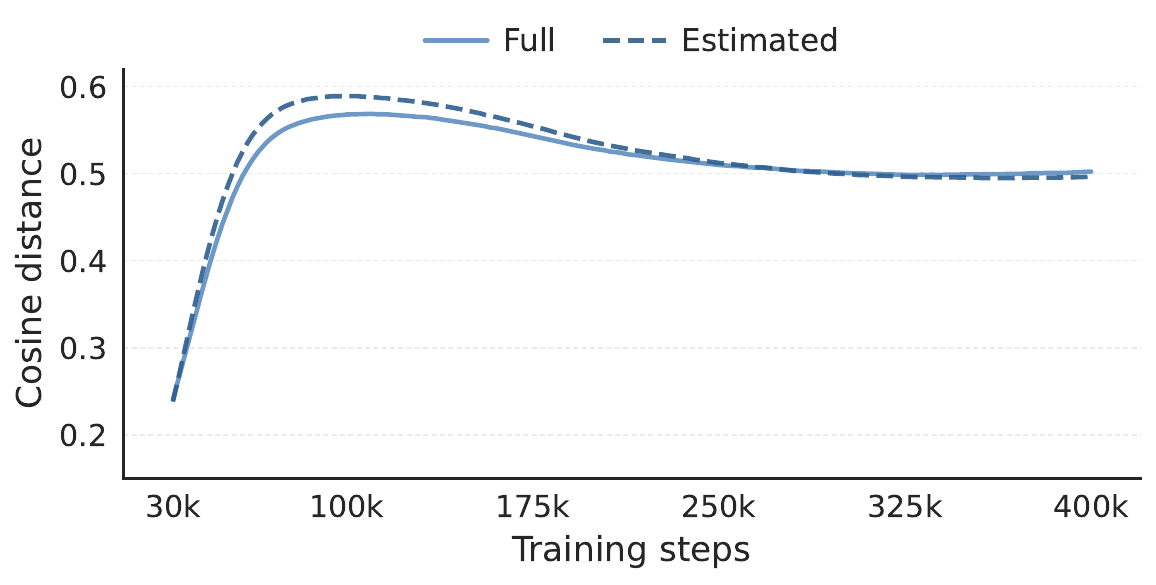}
  \caption{Dense trajectories of the estimated and full $m_{\mathsf{cran}}$ throughout SiT-B/2 training. Curves show 1K-step local averages sampled every 2K steps.}
  \label{fig:estimated-mcran-accuracy}
\end{figure}

We further compare the estimated and full anchors throughout training. Figure~\ref{fig:estimated-mcran-accuracy} reports 1K-step local averages every 2K steps from 30K to 400K. The estimated trajectory closely tracks the full computation: the mean absolute difference in cosine distance is 0.010 over the dense trajectory, and the final values differ by only 0.006.

Finally, we quantify the saved computation using the same optimized CRAN benchmark as Table~\ref{tab:efficiency-profile}. We change only the source of $m_{\mathsf{cran}}$ and run the full and estimated variants in two paired timing runs under the same concurrent H200 load. The average measured time ratio is 0.956. Applying this ratio to the separately measured 43.20\,ms full-geometry time gives 41.32\,ms per step, corresponding to a 4.4\% speedup.

\begin{table}[H]
\centering
\small
\caption{SiT-B/2 training cost of the full and estimated clean inter-sample geometry. The estimated time is obtained by scaling the full-geometry time with the paired measured ratio.}
\label{tab:estimated-mcran-speed}
\begin{tabular}{lcc}
\toprule
$m_{\mathsf{cran}}$ computation & Time/step & Relative time \\
\midrule
Full & 43.20 ms & 1.000$\times$ \\
Estimated & \textbf{41.32 ms} & \textbf{0.956$\times$} \\
\bottomrule
\end{tabular}
\end{table}